# Sapa: A Multi-objective Metric Temporal Planner

**Minh B. Do** BINHMINH@ASU.EDU
**Subbarao Kambhampati** RAO@ASU.EDU
*Department of Computer Science and Engineering*
*Arizona State University, Tempe AZ 85287-5406*

## Abstract

*Sapa* is a domain-independent heuristic forward chaining planner that can handle durative actions, metric resource constraints, and deadline goals. It is designed to be capable of handling the multi-objective nature of metric temporal planning. Our technical contributions include (i) planning-graph based methods for deriving heuristics that are sensitive to both cost and makespan (ii) techniques for adjusting the heuristic estimates to take action interactions and metric resource limitations into account and (iii) a linear time greedy post-processing technique to improve execution flexibility of the solution plans. An implementation of *Sapa* using many of the techniques presented in this paper was one of the best domain independent planners for domains with metric and temporal constraints in the third International Planning Competition, held at AIPS-02. We describe the technical details of extracting the heuristics and present an empirical evaluation of the current implementation of *Sapa*.

## 1. Introduction

The success of the Deep Space Remote Agent experiment has demonstrated the promise and importance of metric temporal planning for real-world applications. HSTS/RAX, the planner used in the remote agent experiment, was predicated on the availability of domain- and planner-dependent control knowledge, the collection and maintenance of which is admittedly a laborious and error-prone activity. An obvious question is whether it will be possible to develop *domain-independent* metric temporal planners that are capable of scaling up to such domains. The past experience has not been particularly encouraging. Although there have been some ambitious attempts–including IxTeT (Ghallab & Laruelle, 1994) and Zeno (Penberthy & Well, 1994), their performance has not been particularly satisfactory.

Some encouraging signs however are the recent successes of domain-independent heuristic planning techniques in classical planning (c.f., Nguyen, Kambhampati, & Nigenda, 2001; Bonet, Loerincs, & Geffner, 1997; Hoffmann & Nebel, 2001). Our research is aimed at building on these successes to develop a scalable metric temporal planner. At first blush search control for metric temporal planners would seem to be a very simple matter of adapting the work on heuristic planners in classical planning (Bonet et al., 1997; Nguyen et al., 2001; Hoffmann & Nebel, 2001). The adaptation however does pose several challenges:

- Metric temporal planners tend to have significantly larger search spaces than classical planners. After all, the problem of planning in the presence of durative actions and metric resources subsumes both classical planning and a certain class of scheduling problems.

- Compared to classical planners, which only have to handle the logical constraints between actions, metric temporal planners have to deal with many additional types of constraints that involve time and continuous functions representing different types of resources.
- In contrast to classical planning, where the only objective is to find shortest length plans, metric temporal planning is *multi-objective*. The user may be interested in improving either temporal quality of the plan (e.g. makespan) or its cost (e.g. cumulative action cost, cost of resources consumed etc.), or more generally, a combination thereof. Consequently, effective plan synthesis requires heuristics that are able to track both these aspects in an evolving plan. Things are further complicated by the fact that these aspects are often inter-dependent. For example, it is often possible to find a "cheaper" plan for achieving goals, if we are allowed more time to achieve them.

In this paper, we present *Sapa*, a heuristic metric temporal planner that we are currently developing to address these challenges. *Sapa* is a forward chaining planner, which searches in the space of time-stamped states *Sapa* handles durative actions as well as actions consuming continuous resources. Our main focus has been on the development of heuristics for focusing *Sapa*'s multi-objective search. These heuristics are derived from the optimistic reachability information encoded in the planning graph. Unlike classical planning heuristics (c.f., Nguyen et al., 2001)), which need only estimate the "length" of the plan needed to achieve a set of goals, *Sapa*'s heuristics need to be sensitive to both the cost and length ("makespan") of the plans for achieving the goals. Our contributions include:

- We present a novel framework for tracking the cost of literals (goals) as a function of time. These "cost functions" are then used to derive heuristics that are capable of directing the search towards plans that satisfy any type of cost-makespan tradeoffs.
- *Sapa* generalizes the notion of "phased" relaxation used in deriving heuristics in planners such as AltAlt and FF (Nguyen et al., 2001; Hoffmann & Nebel, 2001). Specifically, the heuristics are first derived from a relaxation that ignores the delete effects and metric resource constraints, and are then adjusted subsequently to better account for both negative interactions and resource constraints.
- *Sapa* improves the temporal flexibility of the solution plans by post-processing these plans to produce order constrained (o.c or partially-ordered) plans. This way, *Sapa* is able to exploit both the ease of resource reasoning offered by the position-constrained plans and the execution flexibility offered by the precedence-constrained plans. We present a linear time greedy approach to generate an o.c plan of better or equal makespan value compared to a given p.c plan.

**Architecture of *Sapa*:** Figure 1 shows the high-level architecture of *Sapa*. *Sapa* uses a forward chaining A* search to navigate in the space of time-stamped states. Its evaluation function (the "$f(.)$" function is multi-objective and is sensitive to both makespan and action cost. When a state is picked from the search queue and expanded, *Sapa* computes heuristic estimates of each of the resulting children states. The heuristic estimation of a state $S$ is based on (i) computing a relaxed temporal planning graph (RTPG) from $S$, (ii) propagating cost of achievement of literals in the RTPG with the help of time-sensitive cost functions (iii) extracting a relaxed plan $P^r$ for supporting the goals of the problem and (iv) modifying the structure of $P^r$ to adjust for mutex and resource-based interactions. Finally, $P^r$ is used as the basis for deriving the heuristic estimate of $S$. The search ends when a state $S'$ selected for expansion satisfies the goals. In this case, *Sapa* post-processes the position-constrained plan corresponding to the state $S$ to convert it into an order constrained plan. This last step is done to improve the makespan as well as the execution flexibility of the solution plan.

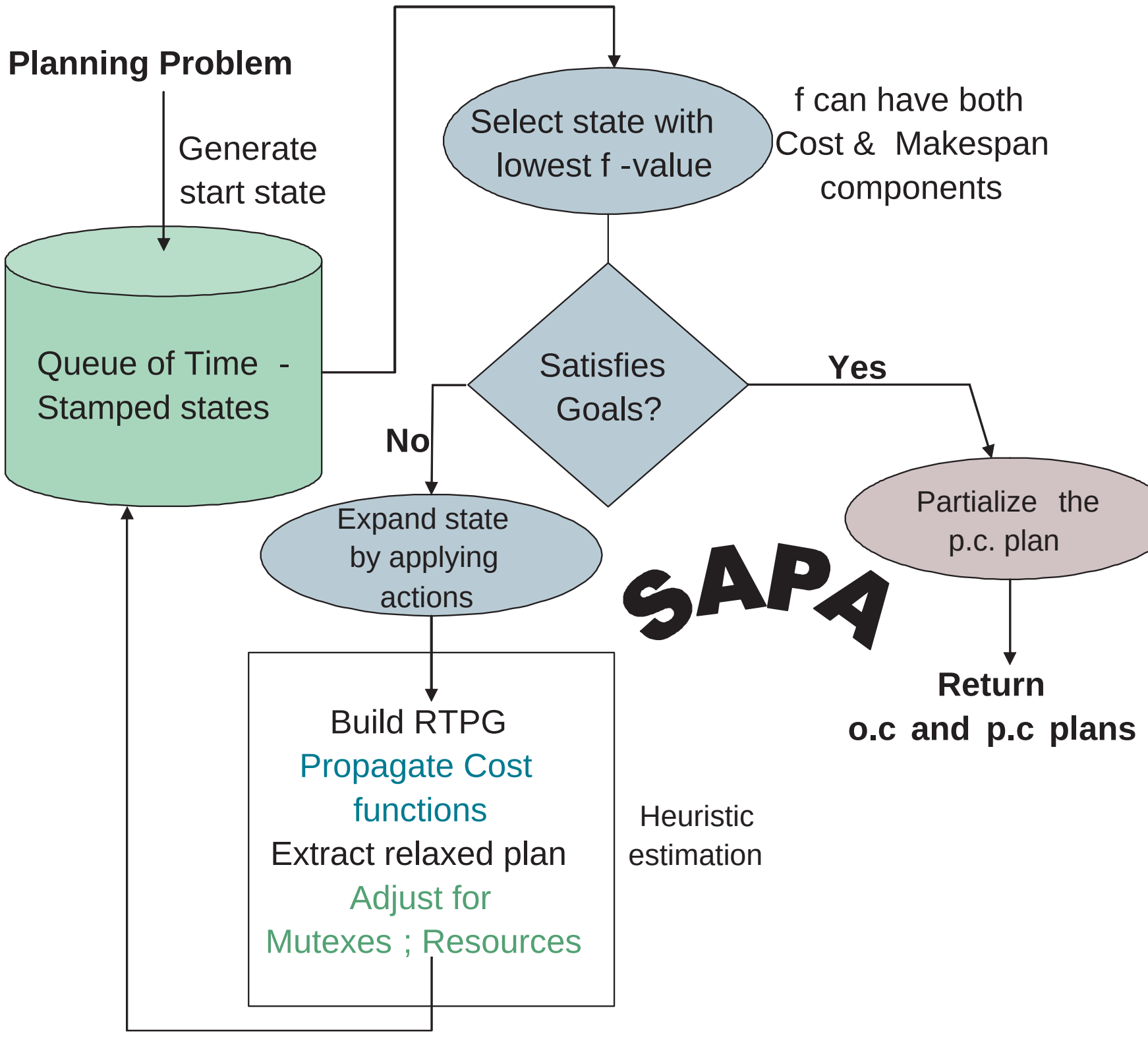

Figure 1: Architecture of *Sapa*

A version of *Sapa* using a subset of the techniques discussed in this paper performed well in domains with metric and temporal constraints in the third International Planning Competition, held at AIPS 2002 (Fox & Long, 2002). In fact, it is the best planner in terms of solution quality and number of problems solved in the highest level of PDDL2.1 used in the competition for the domains *Satellite* and *Rovers*. These domains are both inspired by NASA applications.

**Organization:** The paper is organized as follows: in Section 2 we discuss the details of the action and problem representation, and the forward state space search algorithm used to produce concurrent metric temporal plans with durative actions. In Section 3, we address the problem of propagating the time and cost information over a temporal planning graph. Section 4 shows how the propagated information can be used to estimate the cost of achieving the goals from a given state. We also discuss in that section how the mutual exclusion relations and resource information help improve the heuristic estimation. To improve the quality of the solution, Section 6 discusses our greedy approach of building a precedence-constrained plan from the position-constrained plan returned by *Sapa*. Section 7 discusses the implementation of *Sapa*, presents some empirical results where *Sapa* produces plans with tradeoffs between cost and makespan, and analyzes its performance in the 2002 International Planning Competition (IPC 2002). We present a discussion of the related work in Section 9 and conclude in Section 10.

## 2. Handling Concurrent Actions in a Forward State Space Planner

*Sapa* addresses planning problems that involve durative actions, metric resources, and deadline goals. In this section, we describe how such planning problems are represented and solved in *Sapa*. We first describe the action representation, and then present the forward chaining state search algorithm used by *Sapa*.

### 2.1 Action Representation & Constraints

Planning is the problem of finding a set of actions and the start times of their execution to satisfy all causal, metric, and resource constraints. In this section, we will briefly describe our representation, which is an extension of the action representation in PDDL2.1 Level 3 (Fox & Long, 2001), the most expressive representation level used in the third international competition. Our extensions to PDDL2.1 are: (i) interval preconditions; (ii) delayed effects that happen at time points other than action's start and end time points; (iii) deadline goals.

We shall start with an example to illustrate the action representation in a simple temporal planning problem. This problem and its variations will be used as the running examples throughout the rest of the paper. Figure 2 shows graphically the problem description. In this problem, a group of students in Tucson need to go to Los Angeles (LA). There are two car rental options. If the students rent a faster but more expensive car (*Car1*), they can only go to Phoenix (PHX) or Las Vegas (LV). However, if they decide to rent a slower but cheaper *Car2*, then they can use it to drive to Phoenix or directly to LA. Moreover, to reach LA, the students can also take a *train* from LV or a flight from PHX. In total, there are 6 movement actions in the domain: *drive-car1-tucson-phoenix* ($D^{c_1}_{t\rightarrow p}$, *Dur = 1.0, Cost = 2.0*), *drive-car1-tucson-lv* ($D^{c_1}_{t\rightarrow lv}$, *Dur = 3.5, Cost = 3.0*), *drive-car2-tucson-phoenix* ($D^{c_2}_{t\rightarrow p}$, *Dur = 1.5, Cost = 1.5*), *drive-car2-tucson-la* ($D^{c_2}_{t\rightarrow la}$),*Dur = 7.0, Cost = 6.0, fly-airplane-phoenix-la* ($F_{p\rightarrow la}$, *Dur = 1.5, Cost = 6.0*), and *use-train-lv-la* ($T_{lv\rightarrow la}$, *Dur = 2.5, Cost = 2.5*). Each move action $A$ (by car/airplane/train) between two cities $X$ and $Y$ requires the precondition that the students be at $X$ ($at(X)$) at the beginning of $A$. There are also two temporal effects: $\neg at(X)$ occurs at the starting time point of $A$ and $at(Y)$ at the end time point of $A$. Driving and flying actions also consume different types of resources (e.g fuel) at different rates depending on the specific car or airplane used. In addition, there are refueling actions for cars and airplanes. The durations of the refueling actions depend on the amount of fuel remaining in the vehicle and the refueling rate. The summaries of action specifications for this example are shown on the right side of Figure 2. In this example, the costs of moving by train or airplane are the respective ticket prices, and the costs of moving by rental cars include the rental fees and gas (resource) costs.

As illustrated in the example, unlike actions in classical planning, in planning problems with temporal and resource constraints, actions are not instantaneous but have durations. Each action *A* has a duration $D_A$, starting time $S_A$, and end time ($E_A = S_A + D_A$). The value of $D_A$ can be statically defined for a domain, statically defined for a particular planning problem, or can be dynamically decided at the time of execution. For example, in the traveling domain discussed above, boarding a passenger always takes 10 minutes for all problems in this domain. Duration of the action of flying an airplane between two cities depends on the distance between these two cities and the speed of the airplane. Because the distance between two cities will not change over time, the duration of a particular flying action will be totally specified once we parse the planning problem. However, *refueling* an airplane has a duration that depends on the current fuel level of that airplane. We may only be able to calculate the duration of a given *refueling* action according to the fuel level at the exact time instant when the action will be executed.

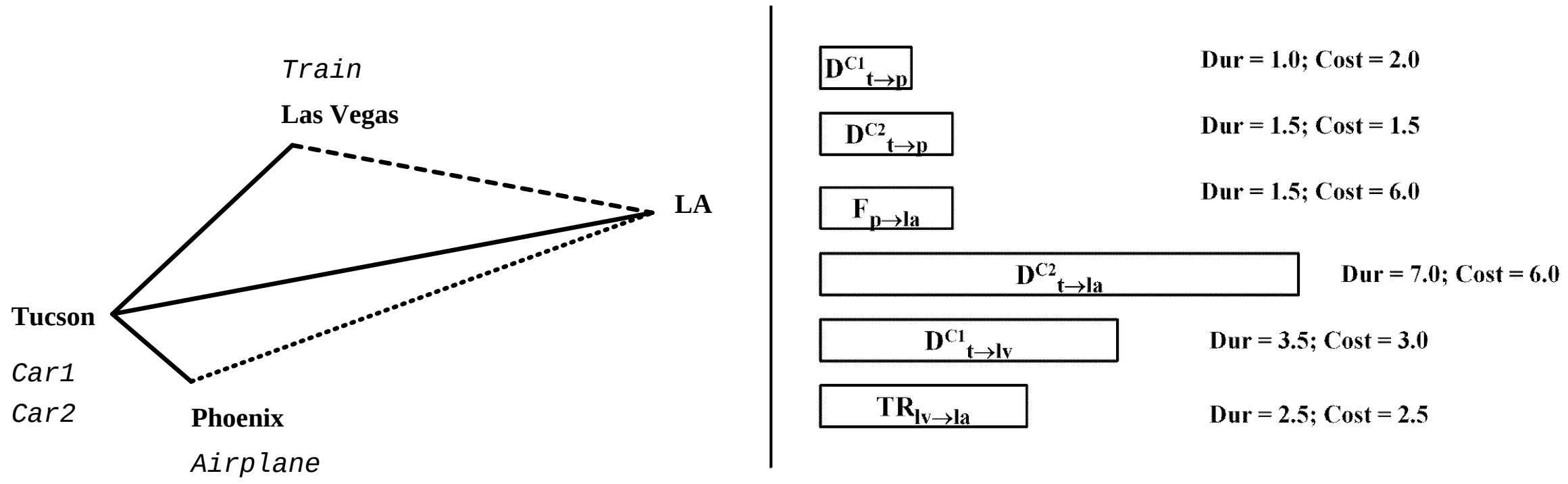

Figure 2: The travel example

An action *A* can have preconditions $Pre(A)$ that may be required either to be instantaneously true at the time point $S_A$ or $E_A$, or required to be true starting at $S_A$ and remain true for some duration $d \leq D_A$. The logical effects *Eff(A)* of *A* are divided into two sets $E_s(A)$, and $E_d(A)$ containing, respectively, the instantaneous effects at time points $S_A$, and delayed effects at $S_A + d, d \leq D_A$. In PDDL2.1, $d$ must be equal to $D_A$ for durative preconditions and delayed effects.

Actions can also consume or produce metric resources and their preconditions may also depend on the values of those resources. For resource related preconditions, we allow several types of equality or inequality checking including ==, <, >, <=, >=. For resource-related effects, we allow the following types of change (update): assignment(=), increment(+=), decrement(-=), multiplication(*=), and division(/=). In essence, actions consume and produce metric resources in the same way that is specified in PDDL2.1.

## 2.2 A Forward Chaining Search Algorithm for metric temporal planning

Variations of the action representation scheme described in the previous section have been used in partial order temporal planners such as IxTeT (Ghallab & Laruelle, 1994) and Zeno (Penberthy & Well, 1994). Bacchus and Ady (2001) were the first to propose a forward chaining algorithm capable of using this type of action representation and still allow concurrent execution of actions in the plan. We adopt and generalize their search algorithm in *Sapa*. The main idea here is to separate the decisions of "which action to apply" and "at what time point to apply the action." Regular progression search planners apply an action in the state resulting from the application of all the actions in the current prefix plan. This means that the start time of the new action is *after* the end time of the last action in the prefix, and the resulting plan will not allow concurrent execution. In contrast, *Sapa* non-deterministically considers (a) application of new actions at the current time stamp (where presumably other actions have already been applied; thus allowing concurrency) and (b) advancement of the current time stamp.

*Sapa*'s search is thus conducted through the space of time stamped states. We define a time stamped state $S$ as a tuple $S = (P, M, \Pi, Q, t)$ consisting of the following structure:

- $P = (\langle p_i, t_i \rangle \mid t_i \leq t)$ is a set of predicates $p_i$ that are true at *t* and $t_i$ is the last time instant at which they were achieved.
- *M* is a set of values for all continuous functions, which may change over the course of planning. Functions are used to represent the metric-resources and other continuous values. Examples of functions are the fuel levels of vehicles.
- $\Pi$ is a set of persistent conditions, such as durative preconditions, that need to be protected during a specific period of time.
- *Q* is an event queue containing a set of updates each scheduled to occur at a specified time in the future. An event *e* can do one of three things: (1) change the True/False value of some predicate, (2) update the value of some function representing a metric-resource, or (3) end the persistence of some condition.
- $t$ is the time stamp of $S$

In this paper, unless noted otherwise, when we say "state" we mean a time stamped state. Note that a time stamped state with a stamp $t$ not only describes the expected snapshot of the world at time $t$ during execution (as done in classical progression planners), but also the delayed (but inevitable) effects of the commitments that have been made by (or before) time $t$.

The initial state $S_{init}$ has time stamp $t = 0$ and has an empty event queue and empty set of persistent conditions. It is completely specified in terms of function and predicate values. The goals are represented by a set of 2-tuples $G = (\langle p_1, t_1 \rangle ... \langle p_n, t_n \rangle)$ where $p_i$ is the $i^{th}$ goal and $t_i$ is the time instant by which $p_i$ needs to be achieved. Note that PDDL2.1 does not allow the specification of goal deadline constraints.

**Goal Satisfaction:** The state $S = (P, M, \Pi, Q, t)$ *subsumes* (entails) the goal *G* if for each $\langle p_i, t_i \rangle \in G$ either:

1. $\exists \langle p_i, t_j \rangle \in P$, $t_j < t_i$ and there is no event in *Q* that deletes $p_i$.
2. $\exists e \in Q$ that adds $p_i$ at time instant $t_e < t_i$, and there is no event in *Q* that deletes $p_i$.[1]

**Action Applicability:** An action A is *applicable* in state $S = (P, M, \Pi, Q, t)$ if:

1. All logical (pre)conditions of *A* are satisfied by *P*.
2. All metric resource (pre)conditions of $A$ are satisfied by *M*. (For example, if the condition to execute an action $A = move(truck, A, B)$ is $fuel(truck) > 500$ then $A$ is executable in $S$ if the value $v$ of $fuel(truck)$ in $M$ satisfies $v > 500$.)

1. In practice, conflicting events are never put on *Q*

3. *A*'s effects do not interfere with any persistent condition in Π and any event in *Q*.

4. There is no event in $Q$ that interferes with persistent preconditions of *A*.

***Interference*:** Interference is defined as the violation of any of the following conditions:

1. Action $A$ should not add any event $e$ that causes $p$ if there is another event currently in $Q$ that causes $\neg p$. Thus, there is never a state in which there are two events in the event queue that cause opposite effects.

2. If $A$ deletes $p$ and $p$ is protected in Π until time point $t_p$, then $A$ should not delete $p$ before $t_p$.

3. If $A$ has a persistent precondition $p$, and there is an event that gives $\neg p$, then that event should occur *after* $A$ terminates.

4. $A$ should not change the value of any function which is currently accessed by another unterminated action[2]. Moreover, $A$ also should not access the value of any function that is currently changed by an unterminated action.

At first glance, the first interference condition seems to be overly strong. However, we argue that it is necessary to keep underlying processes that cause contradicting state changes from overlapping each other. For example, suppose that we have two actions $A_1 = build_house$, $A_2 = destroy_house$ and $Dur(A_1) = 10$, $Dur(A_2) = 7$. $A_1$ has effect $has_house$ and $A_2$ has effect $\neg has_house$ at their end time points. Assuming that $A_1$ is applied at time $t = 0$ and added an event $e = Add(has_house)$ at $t = 10$. If we are allowed to apply $A_2$ at time $t = 0$ and add a contradicting event $e' = Delete(has_house)$ at $t = 7$, then it is unreasonable to believe that we will still have a house at time $t = 10$ anymore. Thus, even though in our current action modeling, state changes that cause $has_house$ and $\neg has_house$ look as if they happen instantaneously at the actions' end time points, there are underlying processes (build/destroy house) that span the whole action durations to make them happen. To prevent those contradicting processes from overlapping with each other, we employ the conservative approach of not letting $Q$ contain contradicting effects.[3]

When we apply an action *A* to a state $S = (P, M, \Pi, Q, t)$, all instantaneous effects of *A* will be immediately used to update the predicate list *P* and metric resources database *M* of *S*. *A*'s persistent preconditions and delayed effects will be put into the persistent condition set Π and event queue *Q* of *S*.

Besides the normal actions, we will have a special action called **advance-time** which we use to advance the time stamp of *S* to the time instant $t_e$ of the earliest event $e$ in the event queue *Q* of *S*. The advance-time action will be applicable in any state *S* that has a non-empty event queue. Upon

2. Unterminated actions are the ones that started before the time point $t$ of the current state $S$ but have not yet finished at $t$.
3. It may be argued that there are cases in which there is no process to give certain effect, or there are situations in which the contradicting processes are allowed to overlap. However, without the ability to explicitly specify the processes and their characteristics in the action representation, we currently decided to go with the conservative approach. We should also mention that the interference relations above *do not* preclude a condition from being established and deleted in the course of a plan as long as the processes involved in establishment and deletion do not overlap. In the example above, it is legal to first build the house and then destroy it.

State Queue: *SQ*={S$_{init}$}
**while** *SQ*≠{}
  *S*:= *Dequeue*(*SQ*)
  **Nondeterministically select** *A* applicable in *S*
    /* $A$ can be advance-time action */
   *S'* := *Apply*(*A*,*S*)
   **if** *S'* satisfies *G* **then** *PrintSolution*
   **else** *Enqueue*(*S'*,*SQ*)
**end while**;

Figure 3: Main search algorithm

applying this action, the state *S* gets updated according to all the events in the event queue that are scheduled to occur at $t_e$. Note that we can apply multiple non-interfering actions at a given time point before applying the special advance-time action. This allows for concurrency in the final plan.

**Search algorithm:** The basic algorithm for searching in the space of time stamped states is shown in Figure 3. We proceed by applying each applicable action to the current state and put each resulting state into the sorted queue using the $Enqueue()$ function. The $Dequeue()$ function is used to take out the first state from the state queue. Currently, *Sapa* employs the A* search. Thus, the state queue is sorted according to some heuristic function that measures the difficulty of reaching the goals from the current state. Next several sections of the paper discuss the design of these heuristic functions.

**Example:** To illustrate how different data structures in the search state $S = (P, M, \Pi, Q, t)$ are maintained during search, we will use a (simpler) variation of our ongoing example introduced at the end of Section 2.1. In this variation, we eliminate the route from Tucson to Los Angeles (LA) going through Las Vegas. Moreover, we assume that there are too many students to fit into one car and they had to be divided into two groups. The first group rents the first car, goes to Phoenix (Phx), and then flies to LA. The second group rents the second car and drives directly to LA. Because the trip from Tucson to LA is very long, the second car needs to be refueled before driving. To further make the problem simpler, we eliminate the boarding/un-boarding actions and assume that the students will reach a certain place (e.g. Phoenix) when their means of transportation (e.g. Car1) arrives there. Figure 4 shows graphically the plan and how the search state $S$'s components change as we go forward. In this example, we assume the $refuel(car)$ action refuels each car to a maximum of 20 gallons. $Drive(car, Tucson, Phoenix)$ takes 8 gallons of gas and $Drive(car, Tucson, LA)$ takes 16 gallons. Note that, at time point $t_1$, event $e_1$ increases the fuel level of $car2$ to 20 gallons. However, the immediately following application of action $A_3$ reduces $fuel(car2)$ back to the lower level of 4 gallons.

## 3. Propagating Time-sensitive Cost Functions in a Temporal Planning Graph

In this section, we discuss the issue of deriving heuristics, that are sensitive to both time and cost, to guide *Sapa*'s search algorithm. An important challenge in finding heuristics to support multi-objective search, as illustrated by the example below, is that the cost and temporal aspects of a plan are often inter-dependent. Therefore, in this section, we introduce the approach of tracking the costs of achieving goals and executing actions in the plan as functions of time. The propagated cost functions can then be used to derive the heuristic values to guide the search in *Sapa*.

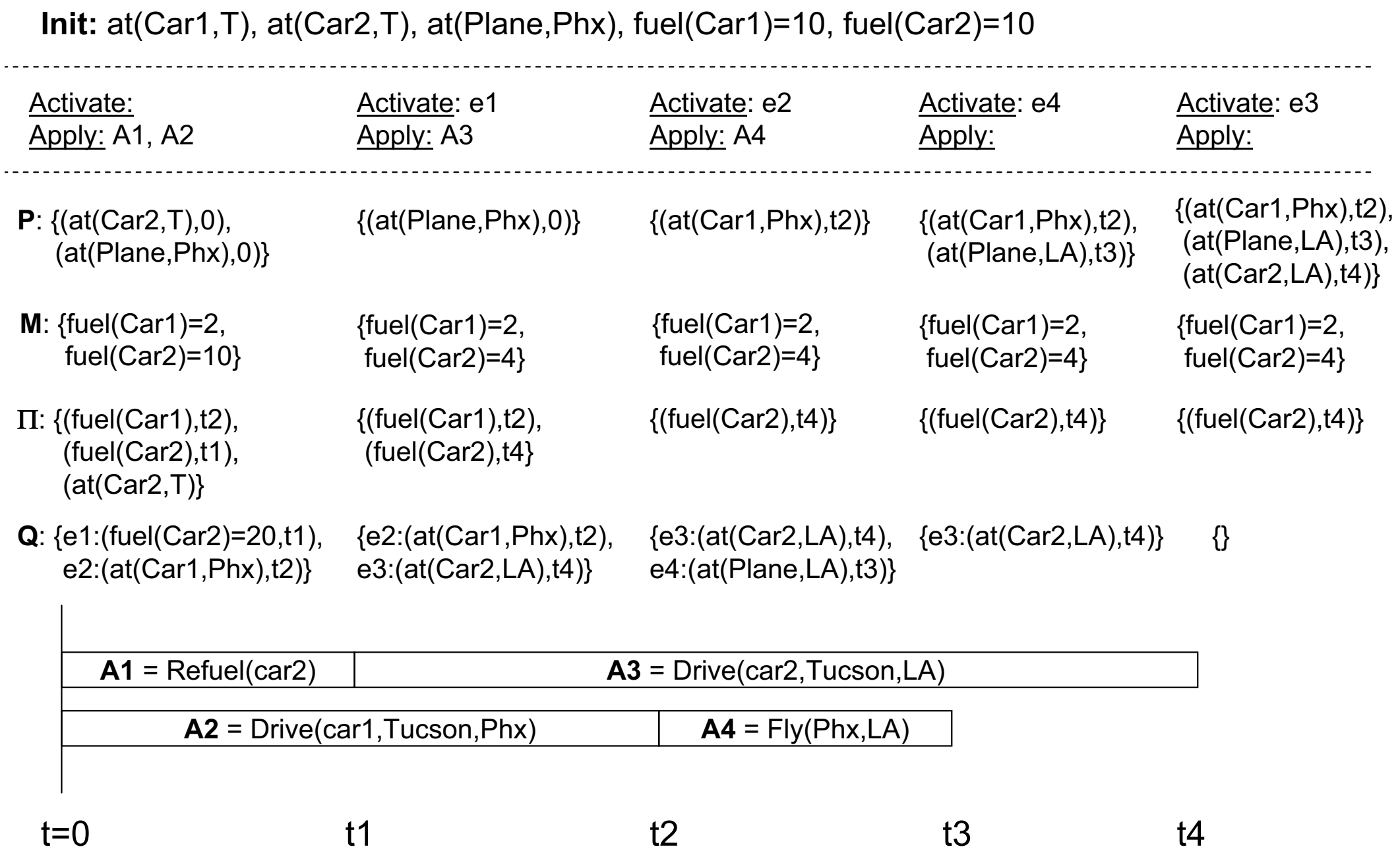

Figure 4: An example showing how different datastructures representing the search state $S = (P, M, \Pi, Q)$ change as we advance the time stamp, apply actions and activate events. The top row shows the initial state. The second row shows the events and actions that are activated and executed at each given time point. The lower rows show how the search state $S = (P, M, \Pi, Q)$ changes due to action application. Finally, we show graphically the durative actions in this plan.

**Example:** Consider a simpler version of our ongoing example. Suppose that we need to go from Tucson to Los Angeles and have two transport options: (i) rent a car and drive from Tucson to Los Angeles in one day for \$100 or (ii) take a shuttle to the Phoenix airport and fly to Los Angeles in 3 hours for \$200. The first option takes more time (higher makespan) but less money, while the second one clearly takes less time but is more expensive. Depending on the specific weights the user gives to each criterion, she may prefer the first option over the second or *vice versa*. Moreover, the user's decision may also be influenced by other constraints on time and cost that are imposed on the final plan. For example, if she needs to be in Los Angeles in six hours, then she may be forced to choose the second option. However, if she has plenty of time but limited budget, then she may choose the first option.

The simple example above shows that makespan and execution cost, while nominally independent of each other, are nevertheless related in terms of the overall objectives of the user and the constraints on a given planning problem. More specifically, for a given makespan threshold (e.g. to be in LA within six hours), there is a certain estimated solution cost tied to it (shuttle fee and ticket price to LA) and analogously for a given cost threshold there is a certain estimated time tied to it. Thus, in order to find plans that are good with respect to both cost and makespan, we need to develop heuristics that track cost of a set of (sub)goals as a function of time.

Given that the planning graph is an excellent structure to represent the relation between facts and actions (c.f., Nguyen et al., 2001), we will use a temporal version of the planning graph structure, such as that introduced in TGP (Smith & Weld, 1999), as a substrate for propagating the cost information. In Section 3.1, we start with a brief discussion of the data structures used for the cost propagation process. We then continue with the details of the propagation process in Section 3.2, and the criteria used to terminate the propagation in Section 3.3.

### 3.1 The Temporal Planning Graph Structure

We now adapt the notion of temporal planning graphs, introduced by Smith and Weld (1999), to our action representation. The temporal planning graph for a given problem is a bi-level graph, with one level containing all *facts*, and the other containing all *actions* in the planning problem. Each fact has links to all actions supporting it, and each action has links to all facts that belong to its precondition and effect lists.[4] Actions are durative and their effects are represented as events that occur at some time between the action's start and end time points. As we will see in more detail in the later parts of this section, we build the temporal planning graph by incrementally increasing the time (makespan value) of the graph. At a given time point $t$, an action $A$ is activated if all preconditions of $A$ can be achieved at $t$. To support the *delayed effects* of the activated actions (i.e., effects that occur at the *future* time points beyond $t$), we also maintain a global event queue for the entire graph, $\mathcal{Q} = \{e_1, e_2, ...e_n\}$ sorted in the increasing order of event time. The event queue for the temporal graph differs from the event queue for the search state (discussed in the previous section) in the following ways:

- It is associated with the whole planning graph (rather than with each single state).
- It only contains the *positive* events. Specifically, the negative effects and the resource-related effects of the actions are not entered in to the graph's queue.
- All the events in $\mathcal{Q}$ have *event costs* associated with each individual event (see below).

Each event in $\mathcal{Q}$ is a 4-tuple $e = \langle f, t, c, A \rangle$ in which: (1) $f$ is the fact that $e$ will add; (2) $t$ is the time point at which the event will occur; and (3) $c$ is the cost incurred to enable the execution of action $A$ which causes $e$. For each action $A$, we introduce a cost function $C(A, t) = v$ to specify the estimated cost $v$ that we incur to enable $A$'s execution at time point $t$. In other words, $C(A, t)$ is the estimate of the cost incurred to achieve all of $A$'s preconditions at time point $t$. Moreover, each action will also have an *execution cost* ($C_{exec}(A)$), which is the cost incurred in executing $A$

4. The bi-level representation has been used in classical planning to save time and space (Long & Fox, 1998), but as Smith & Weld (1999) show, it makes even more sense in temporal planning domains because there is actually no notion of level. All we have are a set of fact/action nodes, each one encoding information such as the *earliest time point* at which the fact/action can be achieved/executed, and the *lowest cost* incurred to achieve them.

**Function** *Propagate Cost*
  Current time: $t_c = 0$;
  $Apply(A_{init}, 0)$;
  **while** *Termination-Criteria* $\neq$ *true*
    Get earliest event $e = \langle f_e, t_e, c_e, A_e \rangle$ from $\mathcal{Q}$;
    $t_c = t_e$;
    **if** $c_e < C(f, t_c)$ **then**
    Update: $C(f, t) = c_e$
      **for all** action $A$: $f \in Precondition(A)$
        $NewCost_A = CostAggregate(A, t_c)$;
        **if** $NewCost_A < C(A, t_c)$ **then**
          Update: $C(A, t) = NewCost(A), t_c \leq t < \infty$;
          $Apply(A, t_c)$;
**End** *Propagate Cost*;

**Function** $Apply(A, t)$
  **For all** A's effect that add $f$ at $S_A + d$ **do**
    $\mathcal{Q} = \mathcal{Q} \bigcup \{e = \langle f, t + d, C(A, t) + C_{exec}(A), A \rangle\}$;
**End** $Apply(A, t)$;

Figure 5: Main cost propagation algorithm

(e.g. ticket price for the *fly* action, gas cost for driving a car). For each fact $f$, a similar cost function $C(f, t) = v$ specifies the estimated cost $v$ incurred to achieve $f$ at time point $t$ (e.g. cost incurred to be in Los Angeles in 6 hours). We also need an additional function $SA(f, t) = A_f$ to specify the action $A_f$ that can be used to *support* $f$ with cost $v$ at time point $t$.

Since we are using a "relaxed" planning graph that is constructed ignoring delete effects, and resource effects, the derived heuristics will not be sensitive to negative interactions and resource restrictions. In Sections 5.1 and 5.2 we discuss how the heuristic measures are adjusted to take these interactions into account.

### 3.2 Cost Propagation Procedure

As mentioned above, our general approach is to propagate the estimated costs incurred to achieve facts and actions from the initial state. As a first step, we need to initialize the cost functions $C(A, t)$ and $C(f, t)$ for all facts and actions. For a given initial state $S_{init}$, let $F = \{f_1, f_2 ... f_n\}$ be the set of facts that are true at time point $t_{init}$ and $\{(f'_1, t_1), ...(f'_m, t_m)\}$, be a set of outstanding positive events which specify the addition of facts $f'_i$ at time points $t_i > t_{init}$. We introduce a dummy action $A_{init}$ to represent $S_{init}$ where $A_{init}$ (i) requires no preconditions; (ii) has cost $C_{exec}(A_{init}) = 0$ and (iii) causes the events of adding all $f_i$ at $t_{init}$ and $f'_i$ at time points $t_i$. At the beginning ($t = 0$), the event queue $\mathcal{Q}$ is empty, the cost functions for all facts and actions are initialized as: $C(A, t) = \infty, C(f, t) = \infty$, $\forall 0 \leq t < \infty$, and $A_{init}$ is the only action that is applicable.

Figure 5 summarizes the steps in the cost propagation algorithm. The main algorithm contains two interleaved parts: one for applying an action and the other for activating an event representing the action's effect.

**Action Introduction:** When an action $A$ is introduced into the planning graph, we (1) augment the event queue $\mathcal{Q}$ with events corresponding to all of $A$'s effects, and (2) update the cost function $C(A,t)$ of $A$.

**Event Activation:** When an event $e = \langle f_e, t_e, C_e, A_e \rangle \in \mathcal{Q}$, which represents an effect of $A_e$ occurring at time point $t_e$ and adding a fact $f_e$ with cost $C_e$ is activated, the cost function of the fact $f_e$ is updated if $C_e < C(f_e, t_e)$. Moreover, if the newly improved cost of $f_e$ leads to a reduction in the cost function of any action $A$ that $f_e$ supports (as decided by function $CostAggregate(A,t)$ in line 11 of Figure 5) then we will ***(re)apply*** $A$ in the graph to propagate $f_e$'s new cost of achievement to the cost functions of $A$ and its effects.

At any given time point $t$, $C(A,t)$ is an aggregated cost (returned by function $CostAggregate(A,t)$) to achieve all of its preconditions. The aggregation can be done in different ways:

1. **Max-propagation:**
   $C(A,t) = Max\{C(f,t) : f \in Precond(A)\}$ or

2. **Sum-propagation:**
   $C(A,t) = \sum\{C(f,t) : f \in Precond(A)\}$ or

The first method assumes that all preconditions of an action depend on each other and the cost to achieve all of them is equal to the cost to achieve the costliest one. This rule leads to the underestimation of $C(A,t)$ and the value of $C(A,t)$ is admissible. The second method (*sum-propagation*) assumes that all facts are independent and is thus inadmissible when subgoals have positive interactions. In classical planning scenarios, sum combination has proved to be more effective than the admissible but much less informed max combination (Nguyen et al., 2001; Bonet et al., 1997).

When the cost function of one of the preconditions of a given action is updated (lowered), the $CostAggregate(A,t)$ function is called and it uses one of the methods described above to calculate if the cost required to execute an action has improved (been reduced).[5] If $C(A,t)$ has improved, then we will *re-apply* $A$ (line 12-14 in Figure 5) to propagate the improved cost $C(A,t)$ to the cost functions $C(f,t)$ of its effects.

The only remaining issue in the main algorithm illustrated in Figure 5 is the *termination criteria* for the propagation, which will be discussed in detail in Section 3.3. Notice that the way we update the cost functions of facts and actions in the planning domains described above shows the challenges in heuristic estimation in temporal planning domains. Because an action's effects do not occur instantaneously at the action's starting time, concurrent actions overlap in many possible ways and thus the cost functions, which represent the difficulty of achieving facts and actions are *time-sensitive*.

Before demonstrating the cost propagation process in our ongoing example, we make two observations about our propagated cost function:

**Observation 1:** *The propagated cost functions of facts and actions are non-increasing over time.*

**Observation 2:** *Because we increase time in* steps *by going through events in the event queue, the cost functions for all facts and actions will be step-functions, even though time is measured continuously.*

5. Propagation rule (2) and (3) will improve (lower) the value of $C(A,t)$ when the cost function of one of $A$'s preconditions is improved. However, for rule (1), the value of $C(A,t)$ is improved only when the cost function of its costliest precondition is updated.

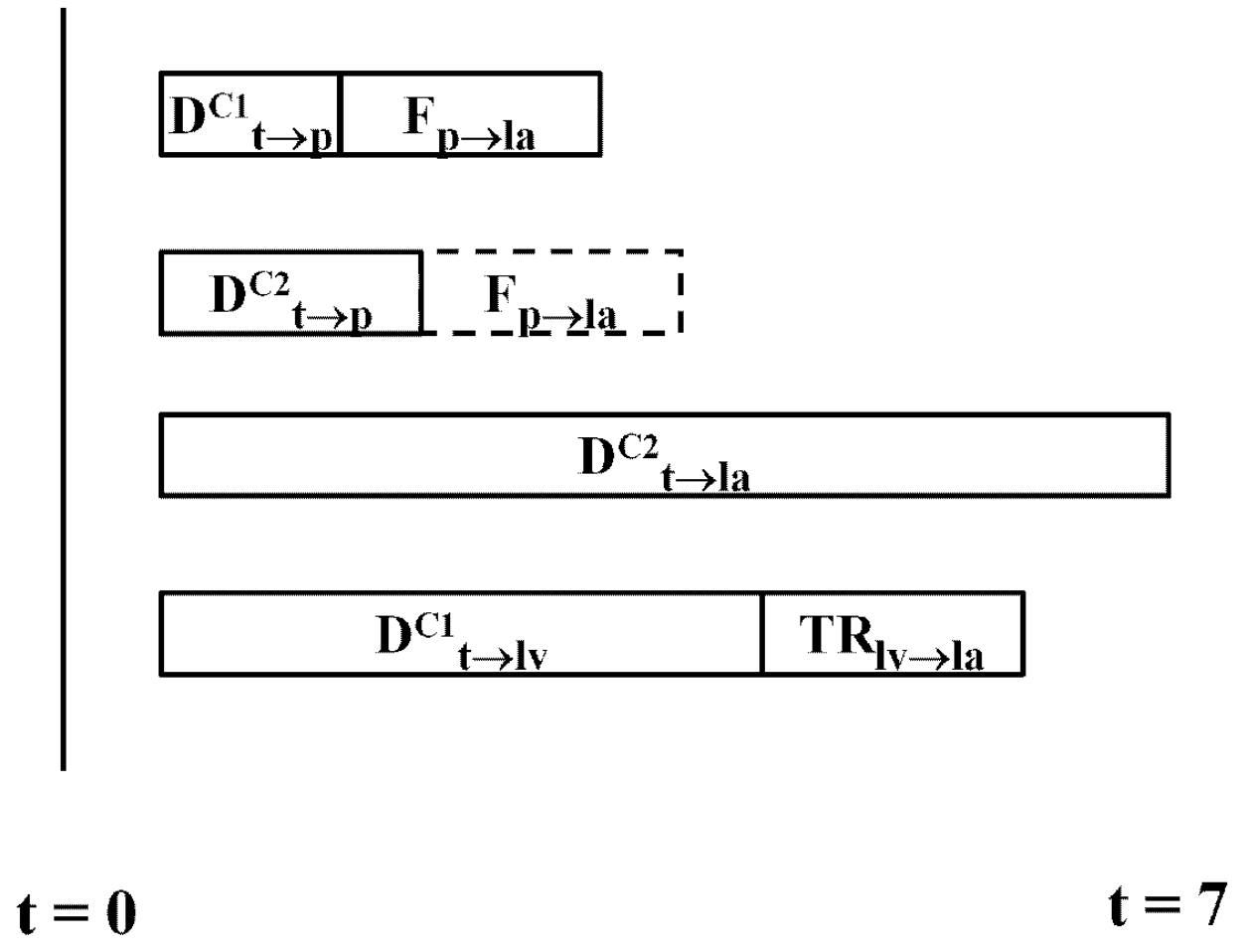

Figure 6: Timeline to represent actions at their earliest possible execution times in the relaxed temporal planning graph.

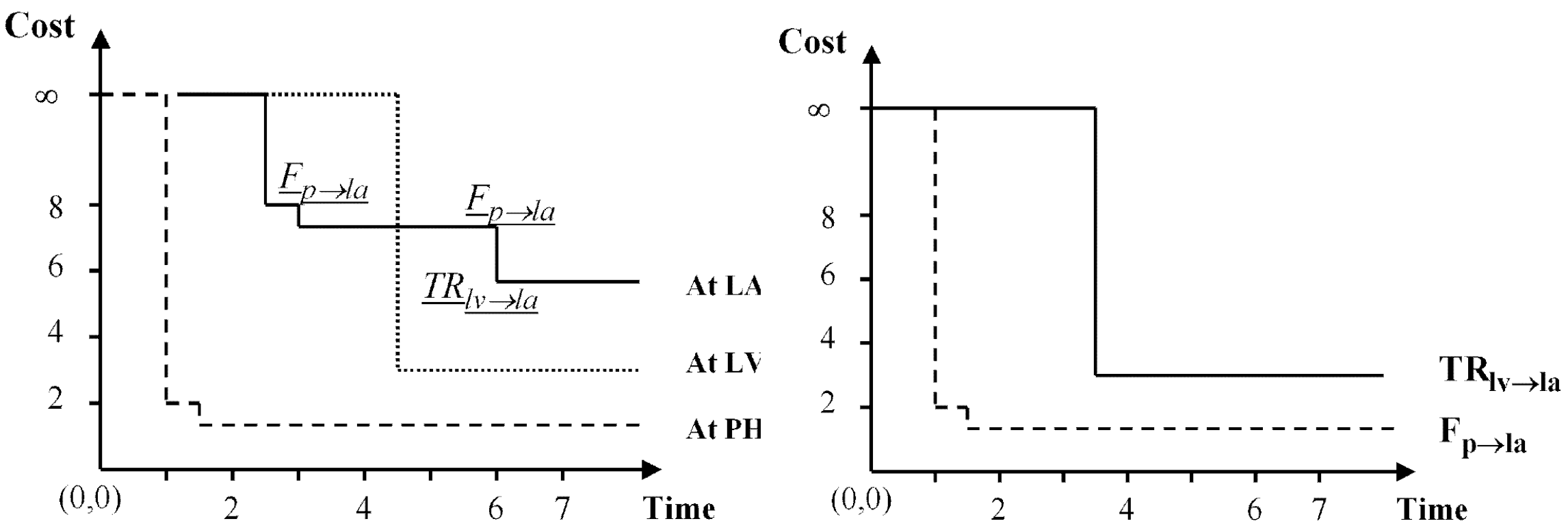

Figure 7: Cost functions for facts and actions in the travel example.

From the first observation, the estimated cheapest cost of achieving a given goal $g$ at time point $t_g$ is $C(g, t_g)$. We do not need to look at the value of $C(g, t)$ at time point $t < t_g$. The second observation helps us in efficiently evaluating the heuristic value for an objective function $f$ involving both time and cost. Specifically, we need compute $f$ at only the (finite number of)time points where the cost function of some fact or action changes. We will come back to the details of the heuristic estimation routines in Section 4.

Returning to our running example, Figure 6 shows graphically the earliest time point at which each action can be applied ($C(A, t) < \infty$) and Figure 7 shows how the cost function of facts/actions change as the time increases. Here is an outline of the update process in this example: at time point $t = 0$, four actions can be applied. They are $D^{c_1}_{t\rightarrow p}$, $D^{c_2}_{t\rightarrow p}$, $D^{c_1}_{t\rightarrow lv}$, $D^{c_2}_{t\rightarrow la}$. These actions add 4 events into the event queue $\mathcal{Q} = \{e_1 = \langle at_phx, t = 1.0, c = 2.0, D^{c_1}_{t\rightarrow p}\rangle$, $e_2 = \langle at_phx, 1.5, 1.5, D^{c_2}_{t\rightarrow p}\rangle$, $e_3 = \langle at_lv, 3.5, 3.0, D^{c_1}_{t\rightarrow lv}\rangle$, $e_4 = \langle at_la, 7.0, 6.0, D^{c_2}_{t\rightarrow la}\rangle\}$. After we advance the time to $t = 1.0$, the first event $e_1$ is activated and $C(at_phx, t)$ is updated. Moreover, because $at_phx$ is a precondition of $F_{p\rightarrow la}$, we also update $C(F_{p\rightarrow la}, t)$ at $t_e = 1.0$ from $\infty$ to 2.0 and put an event $e = \langle at_la, 2.5, 8.0, F_{p\rightarrow la}\rangle$, which represents $F_{p\rightarrow la}$'s effect, into $\mathcal{Q}$. We then go on with the second event $\langle at_phx, 1.5, 1.5, D^{c_2}_{t\rightarrow p}\rangle$ and lower the cost of the fact $at_phx$ and action $F_{p\rightarrow la}$. Event $e = \langle at_la, 3.0, 7.5, F_{p\rightarrow la}\rangle$ is added as a result of the newly improved cost of $F_{p\rightarrow la}$. Continuing the process, we update the cost function of $at_la$ once at time point $t = 2.5$, and again at $t = 3.0$ as the delayed effects of actions $F_{p\rightarrow la}$ occur. At time point $t = 3.5$, we update the cost value of $at_lv$ and action $T_{lv\rightarrow la}$ and introduce the event $e = \langle at_la, 6.0, 5.5, T_{lv\rightarrow la}\rangle$. Notice that the final event $e' = \langle at_la, 7.0, 6.0, D^{c_2}_{t\rightarrow la}\rangle$ representing a delayed effect of the action $D^{c_2}_{t\rightarrow la}$ applied at $t = 0$ will not cause any cost update. This is because the cost function of $at_la$ has been updated to value $c = 5.5 < c_{e'}$ at time $t = 6.0 < t_{e'} = 7.0$.

Besides the values of the cost functions, Figure 7 also shows the supporting actions ($SA(f, t)$, defined in Section 3.1) for the fact (goal) $at_la$. We can see that action $T_{lv\rightarrow la}$ gives the best cost of $C(at_la, t) = 5.5$ for $t \geq 6.0$ and action $F_{p\rightarrow la}$ gives best cost $C(at_la, t) = 7.5$ for $3.0 \leq t < 5.5$ and $C(at_la, t) = 8.0$ for $2.5 \leq t < 3.0$. The right most graph in Figure 7 shows similar cost functions for the actions in this example. We only show the cost functions of actions $T_{lv\rightarrow la}$ and $F_{p\rightarrow la}$ because the other four actions are already applicable at time point $t_{init} = 0$ and thus their cost functions stabilize at 0.

### 3.3 Termination Criteria for the Cost Propagation Process

In this section, we discuss the issue of when we should terminate the cost propagation process. The first thing to note is that cost propagation is in some ways inherently more complex than makespan propagation. For example, once a set of literals enter the planning graph (and are not mutually exclusive), the estimate of the makespan of the shortest plan for achieving them does not change as we continue to expand the planning graph. In contrast, the estimate of the cost of the cheapest plan for achieving them can change until the planning graph levels off. This is why we need to carefully consider the effect of different criteria for stopping the expansion of the planning graph on the accuracy of the cost estimates. The first intuition is that we should not stop the propagation when there exist top level goals for which the cost of achievement is still infinite (unreached goal). On the other hand, given our objective function of finding the cheapest way to achieve the goals, we need not continue the propagation when there is no chance that we can improve the cost of achieving the goals. From those intuitions, following are several rules that can be used to determine when to terminate propagation:

**Deadline termination:** *The propagation should stop at a time point* $t$ *if: (1)* $\forall\ goal\ G : Deadline(G) \leq t$*, or (2)* $\exists\ goal\ G : (Deadline(G) < t) \wedge (C(G,t) = \infty)$*.*

The first rule governs the hard constraints on the goal deadlines. It implies that we should not propagate beyond the latest goal deadline (because any cost estimation beyond that point is useless), or we can not achieve some goal by its deadline.

With the observation that the propagated costs can change only if we still have some events left in the queue that can possibly change the cost functions of a specific propositions, we have the second general termination rule regarding the propagation:

**Fix-point termination:** *The propagation should stop when there are no more events that can decrease the cost of any proposition.*

The second rule is a qualification for reaching the fix-point in which there is no gain on the cost function of any fact or action. It is analogous to the idea of growing the planning graph until it *levels-off* in classical planning.

Stopping the propagation according to the two general rules above leads us to the best (lowest value) achievable cost estimation for all propositions given a specific initial state. However, there are several situations in which we may want to stop the propagation process earlier. First, propagation until the fix-point, where there is no gain on the cost function of any fact or action, would be too costly (c.f., Nguyen et al., 2001). Second, the cost functions of the goals may reach the fix-point long before the full propagation process is terminated according to the general rules discussed above, where the costs of *all* propositions and actions stabilize.

Given the above motivations, we introduce several different criteria to stop the propagation earlier than is entailed by the fix-point computation:

**Zero-lookahead approximation:** *Stop the propagation at the earliest time point* $t$ *where all the goals are reachable (*$C(G,t) < \infty$*).*

**One-lookahead approximation:** *At the earliest time point* $t$ *where all the goals are reachable, execute all the remaining events in the event queue and stop the propagation.*

One-lookahead approximation looks ahead one step in the (future) event queues when one path to achieve all the goals under the relaxed assumption is guaranteed and hopes that executing all those events would explicate some cheaper path to achieve all goals.[6]

Zero and one-lookahead are examples of a more general $k$-lookahead approximation, in which extracting the heuristic value as soon as all the goals are reachable corresponds to *zero-lookahead* and continuing to propagate until the fix-point corresponds to the *infinite (full) lookahead*. The rationale behind the $k$-lookahead approximation is that when all the goals appear, which is an indication that there exists at least one (relaxed) solution, then we will look ahead one or more steps to see if we can achieve some extra improvement in the cost of achieving the goals (and thus lead to a lower cost solution).[7]

6. Note that even if none of those events is directly related to the goals, their executions can still indirectly lead to better (cheaper) path to reach all the goals.
7. For backward planners where we only need to run the propagation one time, infinite-lookahead or higher levels of lookahead may pay off, while in forward planners where we need to evaluate the cost of goals for each single search state, lower values of $k$ may be more appropriate.

Coming back to our travel example, zero-lookahead stops the propagation process at the time point $t = 2.5$ and the goal cost is $C(in_la, 2.5) = 8.0$. The action chain giving that cost is $\{D^{c1}_{t\rightarrow p}, F_{p\rightarrow la}\}$. With one-lookahead, we find the lowest cost for achieving the goal $in_la$ is $C(in_la, 7.0) = 6.0$ and it is given by the action $(D^{c2}_{t\rightarrow la})$. With two-lookahead approximation, the lowest cost for $in_la$ is $C(in_la, 6.0) = 5.5$ and it is achieved by cost propagation through the action set $\{(D^{c1}_{t\rightarrow lv}, T_{lv\rightarrow la})\}$. In this example, two-lookahead has the same effect as the fix-point propagation (infinite lookahead) if the deadline to achieve $in_la$ is later than $t = 6.0$. If it is earlier, say $Deadline(in_la) = 5.5$, then the one-lookahead will have the same effect as the infinite-lookahead option and gives the cost of $C(in_la, 3.0) = 7.5$ for the action chain $\{D^{c2}_{t\rightarrow phx}, F_{phx\rightarrow la}\}$.

## 4. Heuristics based on Propagated Cost Functions

Once the propagation process terminates, the time-sensitive cost functions contain sufficient information to estimate any makespan and cost-based heuristic value of a given state. Specifically, suppose the planning graph is grown from a state $S$. Then the cost functions for the set of goals $G = \{(g_1, t_1), (g_2, t_2)...(g_n, t_n)\}$, $t_i = Deadline(g_i)$ can be used to derive the following estimates:

- The minimum makespan estimate $T(P_S)$ for a plan starting from $S$ is given by the earliest time point $\tau_0$ at which all goals are reached with finite cost $C(g, t) < \infty$.
- The minimum/maximum/summation estimate of slack $Slack(P_S)$ for a plan starting from $S$ is given by the minimum/maximum/summation of the distances between the time point at which each goal first appears in the temporal planning graph and the deadline of that goal.
- The minimum cost estimate, ($C(g, deadline(g))$), of a plan starting from a state $S$ and achieving a set of goals $G$, $C(P_S, \tau_\infty)$, can be computed by aggregating the cost estimates for achieving each of the individual goals at their respective deadlines.[8] Notice that we use $\tau_\infty$ to denote the time point at which the cost propagation process stops. Thus, $\tau_\infty$ is the time point at which the cost functions for all individual goals $C(f, \tau_\infty)$ have their lowest value.
- For each value $t : \tau_0 < t < \tau_\infty$, the cost estimate of a plan $C(P_S, t)$, which can achieve goals within a given makespan limit of $t$, is the aggregation of the values $C(g_i, t)$ of goals $g_i$.

The makespan and the cost estimates of a state can be used as the basis for deriving heuristics. The specific way these estimates are combined to compute the heuristic values does of course depend on what the user's ultimate objective function is. In the general case, the objective would be a function $f(C(P_S), T(P_S))$ involving both the cost ($C(P_S)$) and makespan ($T(P_S)$) values of the plan. Suppose that the objective function is a linear combination of cost and makespan:

$$h(S) = f(C(P_S), T(P_S)) = \alpha.C(P_S) + (1 - \alpha).T(P_S)$$

If the user only cares about the makespan value ($\alpha = 0$), then $h(S) = T(P_S) = \tau_0$. Similarly, if the user only cares about the plan cost ($\alpha = 1$), then $h(S) = C(P_S, \tau_\infty)$. In the more general

8. If we consider $G$ as the set of preconditions for a dummy action that represents the goal state, then we can use any of the propagation rules (max/sum) discussed in Section 3.2 to directly estimate the total cost of achieving the goals from the given initial state.

case, where $0 < \alpha < 1$, then we have to find the time point $t$, $\tau_0 \leq t \leq \tau_\infty$, such that $h_t(S) = f(C(P_S, t), t) = \alpha.C(P_S, t) + (1 - \alpha).t$ has minimum value.[9]

In our ongoing example, given our goal of being in Los Angeles (*at_la*), if $\alpha = 0$, the heuristic value is $h(S) = \tau_0 = 2.5$ which is the earliest time point at which $C(at_la, t) < \infty$. The heuristic value corresponds to the propagation through action chain $(D^{c1}_{t \rightarrow p}, F_{p \rightarrow la})$. If $\alpha = 1$ and $Deadline(At_{LA}) \geq 6.0$, then $h(S) = 5.5$, which is the cheapest cost we can get at time point $\tau_\infty = 6.0$. This heuristic value represents another solution $(D^{c1}_{t \rightarrow lv}, T_{lv \rightarrow la})$. Finally, if $0 < \alpha < 1$, say $\alpha = 0.55$, then the lowest heuristic value $h(S) = \alpha.C(P_S, t) + (1 - \alpha).t$ is $h(S) = 0.55 * 7.5 + 0.45 * 3.0 = 5.47$ at time point $2.5 < t = 3.0 < 6.0$. For $\alpha = 0.55$, this heuristic value $h(S) = 5.47$ corresponds to yet another solution involving driving part way and flying the rest: $(D^{c2}_{t \rightarrow p}, F_{p \rightarrow la})$.

Notice that in the general case where $0 < \alpha < 1$, even though time is measured continuously, we do not need to check every time point $t$: $\tau_0 < t < \tau_\infty$ to find the value where $h(S) = f(C(P_S, t), t)$ is minimal. This is due to the fact that the cost functions for all facts (including goals) are *step functions*. Thus, we only need to compute $h(S)$ at the time points where one of the cost functions $C(g_i, t)$ changes value. In our example above, we only need to calculate values of $h(S)$ at $\tau_0 = 2.5$, $t = 3.0$ and $\tau_\infty = 6.0$ to realize that $h(S)$ has minimum value at time point $t = 3.0$ for $\alpha = 0.55$.

Before we end this section, we note that when there are multiple goals there are several possible ways of computing $C(P_S)$ from the cost functions of the individual goals. This is a consequence of the fact that there are multiple rules to propagate the cost, and there are also interactions between the subgoals. Broadly, there are two different ways to extract the plan costs. We can either directly use the cost functions of the goals to compute $C(P_S)$, or first extract a relaxed plan from the temporal planning graph using the cost functions, and then measure $C(P_S)$ based on the relaxed plan. We discuss these two approaches below.

### 4.1 Directly Using Cost Functions to Estimate $C(P_S)$

After we terminate the propagation using any of the criteria discussed in Section 3.3, let $G = \{(g_1, t_1), (g_2, t_2)...(g_n, t_n)\}$, $t_i = Deadline(g_i)$ be a set of goals and $C_G = \{c_1, ...c_n | c_i = C(g_i, Deadline(g_i)\}$ be their best possible achievement costs. If we consider $G$ as the set of preconditions for a dummy action that represents the goal state, then we can use any of the propagation rules (max/sum) discussed in Section 3.2 to directly estimate the total cost of achieving the goals from the given initial state. Among all the different combinations of the propagation rules and the aggregation rules to compute the total cost of the set of goals $G$, only the *max-max* (max-propagation to update $C(g_i, t)$, and cost of $G$ is the maximum of the values of $C(g_i, Deadline(g_i))$) is admissible. The *sum-sum* rule, which assumes the total independence between all facts, and the other combinations are different options to reflect the dependencies between facts in the planning problem. The tradeoffs between them can only be evaluated empirically.

Goals: $G = \{(g_1, t_1), (g_2, t_2)...(g_n, t_n)\}$
Actions in the relaxed-plan: $RP = \{\}$
Supported facts: $SF = \{f : f \in InitialStateS\}$
**While** $G \neq \emptyset$
    Select the best action $A$ that support $g_1$ at $t_1$
    $RP = RP + A$
    $t_A = t_1 - Dur(A)$
    Update makespan value $T(RP)$ if $t_A < T(RP)$
    **For all** $f \in Effect(A)$ added by $A$ after
        duration $t_f$ from starting point of $A$ **do**
        $SF = SF \bigcup \{(f, t_A + t_f)\}$
    **For all** $f \in Precondition(A)$ s.t $C(f, t_A) > 0$ **do**
        $G = G \bigcup \{(f, t_A)\}$
    **If** $\exists (g_i, t_i) \in G, (g_i, t_j) \in SF : t_j < t_i$ **Then**
        $G = G \setminus \{(g_i, t_i)\}$
**End while;**

Figure 8: Procedure to extract the relaxed plan

### 4.2 Computing Cost from the Relaxed Plan

To take into account the positive interactions between facts in planning problems, we can do a backtrack-free search from the goals to find a relaxed plan. Then, the total execution cost of actions in the relaxed plan and its makespan can be used for the heuristic estimation. Besides providing a possibly better heuristic estimate, work on FF (Hoffmann & Nebel, 2001) shows that actions in the relaxed plan can also be used to effectively focus the search on the branches surrounding the relaxed solution. Moreover, extracting the relaxed solution allows us to use the resource adjustment techniques (to be discussed in Section 5.2) to improve the heuristic estimations. The challenge here is how to use the cost functions to guide the search for the best relaxed plan and we address this below.

The basic idea is to work backwards, finding actions to achieve the goals. When an action is selected, we add its preconditions to the goal list and remove the goals that are achieved by that action. The partial relaxed plan is the plan containing the selected actions and the causal structure between them. When all the remaining goals are satisfied by the initial state $S$, we have the complete relaxed plan and the extraction process is finished. At each stage, an action is selected so that the complete relaxed plan that contains the selected actions is likely to have the lowest estimated objective value $f(P_S, T_S)$. For a given initial state $S$ and the objective function $h(S) = f(C(P_S), T(P_S))$, Figure 8 describes a greedy procedure to find a relaxed plan given the temporal planning graph. First, let $RP$ be the set of actions in the relaxed plan, $SF$ be the set of time-stamped facts $(f_i, t_i)$ that are currently supported , and $G$ be the set of current goals. Thus, $SF$ is the collection of facts supported by the initial state $S$ and the effects of actions in $RP$, and $G$ is the conjunction of top level goals

9. Because $f(C(P_S, t), t)$ estimates the cost of the (cheapest) plan that achieves all goals with the makespan value $T(P_S) = t$, the minimum of $f(C(P_S, t), t)$ ($\tau_0 \leq t \leq \tau_\infty$) estimates the plan $P$ that achieves the goals from state $S$ and $P$ has a smallest value of $f(C(P_S), T(P_S))$. That value would be the heuristic estimation for our objective function of minimizing $f(C(P_S), T(P_S))$.

and the set of preconditions of actions in $RP$ that are not currently supported by facts in $SF$. The estimated heuristic value for the current (partial) relaxed plan and the current goal set is computed as follows: $h(S) = h(RP) + h(G)$ in which $h(RP) = f(C(RP), T(RP))$. For the given set of goals $G$, $h(G) = min_{\tau_0 < t < \tau_\infty}\ f(C(G, t), t)$ is calculated according to the approach discussed in the previous section (Section 4). Finally, $C(RP) = \sum_{A \in RP} C_{exec}(A)$ and $T(RP)$ is the makespan of $RP$, where actions in $RP$ are aligned according to their causal relationship (see below). We will elaborate on this in the example shown later in this section.

At the start, $G$ is the set of top level goals, $RP$ is empty and $SF$ contains facts in the initial state. Thus $C(RP) = 0$, $T(RP) = 0$ and $h(S) = h(G)$. We start the extraction process by backward search for the *least expensive* action $A$ supporting the first goal $g_1$. By least expensive, we mean that $A$ contributes the smallest amount to the objective function $h(S) = h(RP) + h(G)$ if $A$ is added to the current relaxed plan. Specifically, for each action $A$ that supports $g_1$, we calculate the value $h_A(S) = h(RP + A) + h((G \setminus Effect(A)) \bigcup Precond(A))$ which estimates the heuristic value if we add $A$ to the relaxed plan. We then choose the action $A$ that has the smallest $h_A(S)$ value.

When an action $A$ is chosen, we put its preconditions into the current goal list $G$, and its effects into the set of supported facts $SF$. Moreover, we add a precedence constraint between $A$ and the action $A_1$ that has $g_1$ as its precondition so that $A$ gives $g_1$ before the time point at which $A_1$ needs it. Using these ordering relations between actions in $RP$ and the mutex orderings discussed in Section 5.1, we can update the makespan value $T(RP)$ of the current (partial) relaxed plan.

In our ongoing example, suppose that our objective function is $h(S) = f(C(P_S), T(P_S)) = \alpha.C(P_S) + (1 - \alpha).T(P_S)$, with $\alpha = 0.55$ and the infinite-lookahead criterion is used to stop the cost propagation process. When we start extracting the relaxed plan, the initial setting is $G = \{at_la\}$, $RP = \emptyset$ and $SF = \{at_tucson\}$. Among the three actions $D^{c2}_{t \rightarrow la}$, $T_{lv \rightarrow la}$ and $F_{p \rightarrow la}$ that support the goal $at_la$, we choose action $A = F_{p \rightarrow la}$ because if we add it to the relaxed plan $RP$, then the estimated value $h_A(S) = h(RP + A) + h((G \setminus at_la) \bigcup at_phx)$ = $(\alpha * C_{exec}(F_{p \rightarrow la}) + (1 - \alpha) * Dur(F_{p \rightarrow la})) + min_t(f(C(at_phx), t))$ = (0.55*6.0 + 0.45*1.5) + (0.55*1.5 + 0.45*1.5) = 5.475. This turns out to be the smallest among the three actions. After we add $F_{p \rightarrow la}$ to the relaxed plan, we update the goal set to $G = \{at_phx\}$. It is then easy to compare between the two actions $D^{c2}_{t \rightarrow phx}$ and $D^{c1}_{t \rightarrow phx}$ to see that $D^{c2}_{t \rightarrow phx}$ is cheaper for achieving *at-phx* given the value $\alpha = 0.55$. The final cost $C(P_S) = 6.0 + 1.5 = 7.5$ and makespan of $T(P_S) = 1.5 + 1.5 = 3$ of the final relaxed plan can be used as the final heuristic estimation $h(S) = 0.55 * 7.5 + 0.45 * 3 = 5.475$ for the given state.

Notice that in the relaxed-plan extraction procedure, we set the time points for the goal set to be the goal deadlines, instead of the latest time points where the cost functions for the goals stabilized. The reason is that the cost values of facts and actions *monotonically decrease* and the costs are *time-sensitive*. Therefore, the later we set the time points for goals to start searching for the relaxed plan, the better chance we have of getting the low-cost plan, especially when we use the $k$-lookahead approximation approach with $k \neq \infty$. In our ongoing example, if we use the zero-lookahead option to stop the propagation, we find that the smallest cost is $C(in_la) = 8.0$ at $t = 2.5$. If we search back for the relaxed plan with the combination $(in_la, 2.5)$ then we would find a plan $P_1 = (D^{c1}_{t \rightarrow p}, F_{p \rightarrow la})$. However, if we search from the goal deadline, say $t = 7.0$, then we would realize that the lowest cost for the precondition $in_phx$ of $F_{p \rightarrow la}$ at $t = 7.0 - 1.5 = 5.5$ is $C(in_phx, 5.5) = 1.5$ (caused by $D^{c2}_{t \rightarrow p}$ at time point $t = 2.0$) and thus the final plan is $P_2 = (D^{c2}_{t \rightarrow p}, F_{p \rightarrow la})$ which is cheaper than $P_1$.

### 4.3 Origin of Action Costs

In all our preceding discussion of cost-based heuristics, we have implicitly assumed that the individual action costs are specified directly as part of the problem specification. While this is a reasonable assumption, it can also be argued that unlike the duration, the cost of an action is implicitly dependent on what the user is interested in optimizing. For example, suppose, in a transportation domain, the user declares the objective function to be optimized as:[10]

$$4 * TotalTime + 0.005 * TotalFuelUsed$$

without providing any additional explicit information about action costs. It is possible to use the objective function to assess the costs of individual actions (in terms of how much they contribute to the cost portion of the objective). Specifically, the cost each action can be set equal to the amount of fuel used by that action. The $\alpha$ value (for combining cost and makespan) can be set based on the coefficients in the objective function. Of course, this type of "de-compilation" of the objective function into action costs is only possible if the objective function is a linear combination of makespan and resource consumption.

## 5. Adjustments to the Relaxed Plan Heuristic

Until now, the heuristic estimates have been calculated by relaxing certain types of constraints such as negative effects and metric resource consumptions. In this section, we discuss how those contraints can then be used to adjust and improve the final heuristic values.

### 5.1 Improving the Relaxed Plan Heuristic Estimation with Static Mutex Relations

When building the relaxed temporal planning graph (RTPG), we ignored the negative interactions between concurrent actions. We now discuss a way of using the static mutex relations to help improve the heuristic estimation when extracting the relaxed plan. Specifically, our approach involves the following steps:

1. Find the set of static mutex relations between the ground actions in the planning problem based on their negative interactions.[11]

2. When extracting the relaxed plan (Section 4.2), besides the orderings between actions that have causal relationships (i.e one action gives the effect that supports the other action's preconditions), we also post precedence constraints to avoid concurrent execution of actions that are mutex. Specifically, when a new action is added to the relaxed plan, we use the pre-calculated static mutexes to establish ordering between mutually exclusive action pairs so that they can not be executed concurrently. The orderings are selected in such a way that they violate the least number of existing causal links in the relaxed plan.

By using the mutex relations in this way, we can improve the makespan estimation of the relaxed plan, and thus the heuristic estimation. Moreover, in some cases, the mutex relations can also help us detect that the relaxed plan is in fact a valid plan, and thus can lead to the early termination

10. In fact, this was the metric specified for the first problem in the Zeno-Travel domain in IPC 2003.
11. Two actions are static mutex if the delete effects of one action intersect with the preconditions or add effects of the other.

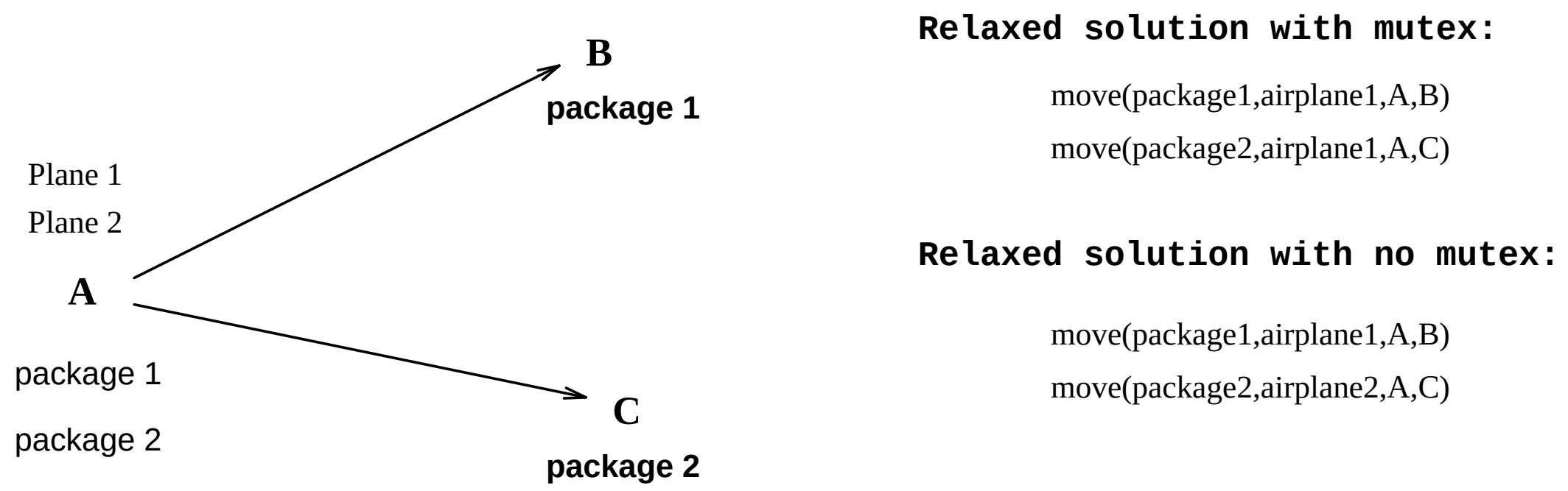

Figure 9: Example of mutex in the relaxed plan

of the search. Consider the example of the Logistics domain illustrated in Figure 9. In this example, we need to move two packages from $cityA$ to $cityB$ and $cityC$ and there are two airplanes ($plane_1, plane_2$) at $cityA$ that can be used to move them. Moreover, we assume that $plane_1$ is 1.5 times faster than $plane_2$ and uses the same amount of resources to fly between two cities. There are two relaxed plans

$$P_1 = \{move(package_1, plane_1, cityA, cityB), move(package_2, plane_1, cityA, cityC)\}$$

$$P_2 = \{move(package_1, plane_1, cityA, cityB), move(package_2, plane_2, cityA, cityC)\}$$

that both contain two actions. The first one uses the same plane to carry both packages, while the second one uses two different planes. The first one has a shorter makespan if mutexes are ignored. However, if we consider the mutex constraints, then we know that two actions in $P_1$ can not be executed concurrently and thus the makespan of $P_1$ is actually longer than $P_2$. Moreover, the static mutex relations also show that even if we order the two actions in $P_1$, there is a violation because the first action cuts off the causal link between the initial state and the second one. Thus, the mutex information helps us in this simple case to find a better (consistent) relaxed plan to use as a heuristic estimate. Here is a sketch of how the relaxed plan $P_2$ can be found. After the first action $A_1 = move(package_1, plane_1, cityA, cityB)$ is selected to support the goal $at(package_1, cityB)$, the relaxed plan is $RP = A_1$ and the two potential actions to support the second goal $at(package_2, cityC)$ are $A_2 = move(package_2, plane_1, cityA, cityC)$ and $A_2' = move(package_2, plane_2, cityA, cityC)$. With mutex information, we will be able to choose $A_2'$ over $A_2$ to include in the final relaxed plan.

### 5.2 Using Resource Information to Adjust the Cost Estimates

The heuristics discussed in Section 4 have used the knowledge about durations of actions and deadline goals but not resource consumption. By ignoring the resource related effects when building the relaxed plan, we may miss counting actions whose only purpose is to provide sufficient resource-related conditions to other actions. In our ongoing example, if we want to drive a car from Tucson to LA and the gas level is low, by totally ignoring the resource related conditions, we will not realize that we need to *refuel* the car before *drive*. Consequently, ignoring resource constraints may reduce the quality of the heuristic estimate based on the relaxed plan. We are thus interested in adjusting the heuristic values discussed in the last two sections to account for the resource constraints.

In many real-world problems, most actions consume resources, while there are special actions that increase the levels of resources. Since checking whether the level of a resource is sufficient for allowing the execution of an action is similar to checking the predicate preconditions, one obvious approach is to adjust the relaxed plan by including actions that provide that resource-related condition to the relaxed plan. However, for many reasons, it turns out to be too difficult to decide which actions should be added to the relaxed plan to satisfy the given resource conditions (Do & Kambhampati, 2001, gives a more detailed discussion of these difficulties). Therefore, we introduce an indirect way of adjusting the cost of the relaxed plan to take into account the resource constraints. We first pre-process the problem specifications and find for each resource *R* an action $A_R$ that can increase the amount of *R* maximally. Let $\Delta_R$ be the amount by which $A_R$ increases *R*, and let $C(A_R)$ be the cost value of $A_R$. Let $Init(R)$ be the level of resource *R* at the state *S* for which we want to compute the relaxed plan, and $Con(R)$, $Pro(R)$ be the total consumption and production of *R* by all actions in the relaxed plan. If $Con(R) > Init(R) + Pro(R)$, then we increase the cost by the number of production actions necessary to make up the difference. More precisely:

$$C \leftarrow C + \sum_R \left\lceil \frac{(Con(R) - (Init(R) + Pro(R)))}{\Delta_R} \right\rceil * C(A_R)$$

We shall call this the adjusted cost heuristic. The basic idea is that even though we do not know if an individual resource-consuming action in the relaxed plan needs another action to support its resource-related preconditions, we can still adjust the number of actions in the relaxed plan by reasoning about the total resource consumption of *all* the actions in the plan. If we know the resources $R$ consumed by the relaxed plan and the maximum production of those resources possible by any individual action in the domain, then we can infer the minimum number of resource-increasing actions that we need to add to the relaxed plan to balance the resource consumption. In our ongoing example, if the car rented by the students at *Tucson* does not have enough fuel in the initial state to make the trip to Phoenix, LA, or Las Vegas, then this approach will discover that the planner needs to add a *refuel* action to the relaxed plan.

Currently, our resource-adjustment technique discussed above is limited to simple consumption and production of resources using addition and subtraction. These are the most common forms, as evidenced by the fact that in all metric temporal planning domains used in the competition, actions consume and produce resources solely using addition (increase) and subtraction (decrease). Modifications are needed to extend our current approach to deal with other types of resource consumption such as using multiplication or division.

## 6. Post-Processing to Improve Temporal Flexibility

To improve the makespan and execution flexibility of the plans generated by *Sapa*, we post-process and convert them into partially ordered plans. We discuss the details of this process in this section. We will start by first differentiating between two broad classes of plans.

**Position and Order constrained plans:** *A position constrained plan (p.c.) is a plan where the execution time of each action is fixed to a specific time point. An order constrained (o.c.) plan is a plan where only the relative orderings between the actions are specified.*

Note that the p.c. vs. o.c. distinction is orthogonal to whether or not concurrency is allowed during execution. Indeed, we can distinguish two subclasses of p.c. plans–serial and parallel. In a serial position constrained plan, no concurrency of execution is allowed. In a parallel position constrained plan, actions are allowed to execute concurrently. Examples of the serial p.c. plans are the ones returned by classical planners such as HSP (Bonet et al., 1997), AltAlt (Nguyen et al., 2001), FF (Hoffmann & Nebel, 2001), or GRT (Refanidis & Vlahavas, 2001b). The parallel p.c. plans are the ones returned by *Sapa* (Do & Kambhampati, 2001), TP4 (Haslum & Geffner, 2001). Graphplan-based planners (STAN, Long & Fox, 1998; IPP, Koehler et al. 1997) and their temporal cousins (TGP, Smith & Weld, 1999; TPSYS, Garrido, Onaindia & Barber, 2001) also return parallel p.c plans. Examples of planners that output order constrained (o.c.) plans are Zeno (Penberthy & Well, 1994), HSTS (Muscettola, 1994), IxTexT (Laborie & Ghallab, 1995), RePOP (Nguyen & Kambhampati, 2001), and VHPOP (Younes & Simmons, 2003).

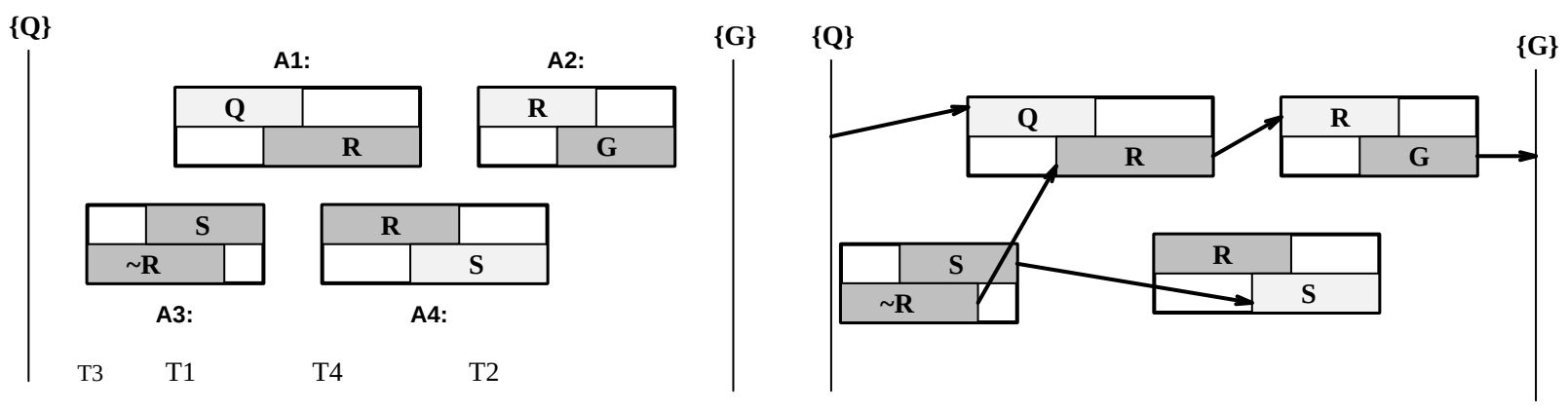

Figure 10: Examples of p.c. and o.c. plans

As mentioned above, the plans returned by *Sapa* are *position-constrained* (p.c.). Searching in the space of these p.c. plans has some advantages in the presence of metric resources (viz., it is easy to compute the amount of consumed resources at every point in a partial plan). Position constrained plans are however less desirable from the point of view of execution flexibility and human comprehension. For these latter uses, an order (precedence) constrained plan (o.c plan) is often better.

Figure 10 shows a valid parallel p.c. plan consisting of four actions $A_1, A_2, A_3, A_4$ with their starting time points fixed to $T_1, T_2, T_3, T_4$ and an o.c plan consisting of the same set of actions and achieving the same goals. For each action, the shaded regions show the durations over which each precondition or effects should hold during each action's execution time. The darker ones represent the effect and the lighter ones represent preconditions. For example, action $A_1$ has a precondition $Q$ and effect $R$; action $A_3$ has no preconditions and two effects $\neg R$ and $S$. The arrows show the relative orderings between actions. Those ordering relations represent the o.c plan and thus any execution trace that does not violate those orderings will be a consistent p.c plan.

Given a p.c plan $P_{pc}$, we are interested in computing an o.c. plan $P_{oc}$ that contains the same actions as $P_{pc}$, and is also a valid plan for solving the original problem. In general, there can be many such o.c. plans. In the related work (Do & Kambhampati, 2003), we discuss how this conversion problem can be posed as an optimization problem subject to any variety of optimization metrics based on temporal quality and flexibility. In the following we discuss a greedy strategy that was used in the competition version of *Sapa*, which finds an o.c plan biased to have a reasonably good makespan. Specifically, we extend the explanation-based order generation method outlined by Kambhampati and Kedar (1994) to first compute a causal explanation for the p.c plan and then construct an o.c plan that has just the number of orderings needed to satisfy that explanation. This strategy depends heavily on the positions of all the actions in the original p.c. plan. It works based on the fact that the alignment of actions in the original p.c. plan guarantees that causation and preservation constraints are satisfied. The following notation helps in describing our approach:

- For each (pre)condition $p$ of action $A$, we use $[st_A^p, et_A^p]$ to represent the duration in which $p$ should hold ($st_A^p = et_A^p$ if $p$ is an instant precondition).
- For each effect $e$ of action $A$, we use $et_A^e$ to represent the time point at which $e$ occurs.
- For each resource $r$ that is checked in preconditions or used by some action $A$, we use $[st_A^r, et_A^r]$ to represent the duration over which $r$ is accessed by $A$.
- The initial and goal states are represented by two new actions $A_I$ and $A_G$. $A_I$ starts before all other actions in the $P_{pc}$, it has no precondition and has effects representing the initial state. $A_G$ starts after all other actions in $P_{pc}$, has no effect, and has top-level goals as its preconditions.
- The symbol $'' \prec''$ is used throughout this section to denote the relative precedence orderings between two time points.

Note that the values of $st_A^p, et_A^p, et_A^e, st_A^r, et_A^r$ are fixed in the p.c plan but are only partially ordered in the o.c plan. The o.c plan $P_{oc}$ is built from a p.c plan $P_{pc}$ as follows:

**Supporting Actions:** For each precondition $p$ of action $A$, we choose the earliest possible action $A'$ in the p.c plan that can support $p$:

1. $p \in \mathit{Effect}(A')$ and $et_{A'}^p < st_A^p$ in the p.c. plan $P_{pc}$.
2. There is no action $B$ such that: $\neg p \in \mathit{Effect}(B)$ and $et_{A'}^p < et_B^{\neg p} < st_A^p$ in $P_{pc}$.
3. There is no other action $C$ that also satisfies the two conditions above and $et_C^p < et_{A'}^p$.

When $A'$ is selected to support $p$ for $A$, we add the causal link $A' \xrightarrow{p} A$ between two time points $et_{A'}^p$ and $st_A^p$ to the o.c plan. Thus, the ordering $et_{A'}^p \prec st_A^p$ is added to $P_{oc}$.

**Interference relations:** For each pair of actions $A, A'$ that interfere with each other, we order them as follows:

1. If $\exists p \in Delete(A') \bigcap Add(A)$, then add the ordering $et_A^p \prec et_{A'}^{\neg p}$ to $P_{oc}$ if $et_A^p < st_{A'}^{\neg p}$ in $P_{pc}$. Otherwise add $et_{A'}^{\neg p} \prec et_A^p$ to $P_{oc}$.
2. If $\exists p \in Delete(A') \bigcap Precond(A)$, then add the ordering $et_A^p \prec et_{A'}^{\neg p}$ to $P_{oc}$ if $et_A^p < et_A^{\neg p}$ in $P_{pc}$. Otherwise, if $et_{A'}^{\neg p} < st_A^p$ in the original plan $P_{pc}$ then we add the ordering $et_{A'}^{\neg p} \prec st_A^p$ to the final plan $P_{oc}$.

**Resource relations:** For each resource $r$ that is checked as a (pre)condition for action $A$ and used by action $A'$, based on those action's fixed starting times in the original p.c plan $P_{pc}$, we add the following orderings to the resulting $P_{oc}$ plan as follows:

- If $et_A^r < st_{A'}^r$ in $P_{pc}$, then the ordering relation $et_A^r \prec st_{A'}^r$ is added to $P_{oc}$.
- If $et_{A'}^r < st_A^r$ in $P_{pc}$, then the ordering relation $et_{A'}^r \prec st_A^r$ is added to $P_{oc}$.

This strategy is backtrack-free due to the fact that the original p.c. plan is correct. All (pre) conditions of all actions in $P_{pc}$ are satisfied and thus for any precondition $p$ of an action $A$, we can always find an action $A'$ that satisfies the three constraints listed above to support $p$. Moreover, one of the temporal constraints that lead to the consistent ordering between two interfering actions (logical or resource interference) will always be satisfied because the p.c. plan is consistent and no pair of interfering actions overlap each other in $P_{pc}$. Thus, the search is backtrack-free and we are guaranteed to find an o.c. plan due to the existence of one legal dispatch of the final o.c. plan $P_{oc}$ (which is the starting p.c. plan $P_{pc}$). The final o.c. plan is valid because there is a causal-link for every action's precondition, all causal links are safe, no interfering actions can overlap, and all the resource-related (pre)conditions are satisfied. Moreover, this strategy ensures that the orderings on $P_{oc}$ are consistent with the original $P_{pc}$. Therefore, because the p.c plan $P_{pc}$ is one among multiple p.c plans that are consistent with the o.c plan $P_{oc}$, the makespan of $P_{oc}$ is guaranteed to be equal or better than the makespan of $P_{pc}$.

The algorithm discussed in this section is a special case of the partialization problem in metric temporal planning. In our related work (Do & Kambhampati, 2003), we do a systematic study of the general partialization problem and give CSOP (Constraint Satisfaction Optimization Problem) encodings for solving them. The current algorithm can be seen as a particular greedy variable/value ordering strategy for the CSOP encoding.

## 7. Implementation of *Sapa*

The *Sapa* system with all the techniques described in this paper has been implemented in Java. The implementation includes:

1. The forward chaining algorithm (Section 2).
2. The cost sensitive temporal planning graph and the routines to propagate the cost information and extract the heuristic value from it (Section 3).
3. The routines to extract and adjust the relaxed plan using static mutex and resource information (Section 4.2).
4. Greedy post-processing routines to convert the p.c. plan into an o.c plan with better makespan and execution flexibility (Section 6).

By default *Sapa* uses the sum-propagation rule, infinite lookahead termination, resource-adjusted heuristics, and converts the solutions into o.c. plans. Besides the techniques described in this paper, we also wrote a JAVA-based parser for PDDL2.1 Level 3, which is the highest level used in the Third International Planning Competition (IPC3).

To visualize the plans returned by *Sapa* and the relations between actions in the plan (such as causal links, mutual exclusions, and resource relations), we have developed a Graphical User Interface (GUI)[12] for *Sapa*. Figure 11 and 12 shows the screen shots of the current GUI. It displays the time line of the final plan with each action shown with its actual duration and starting time in the final plan. There are options to display the causal relations between actions (found using the greedy approach discussed in Section 6), and logical and resource mutexes between actions. The specific times at which individual goals are achieved are also displayed.

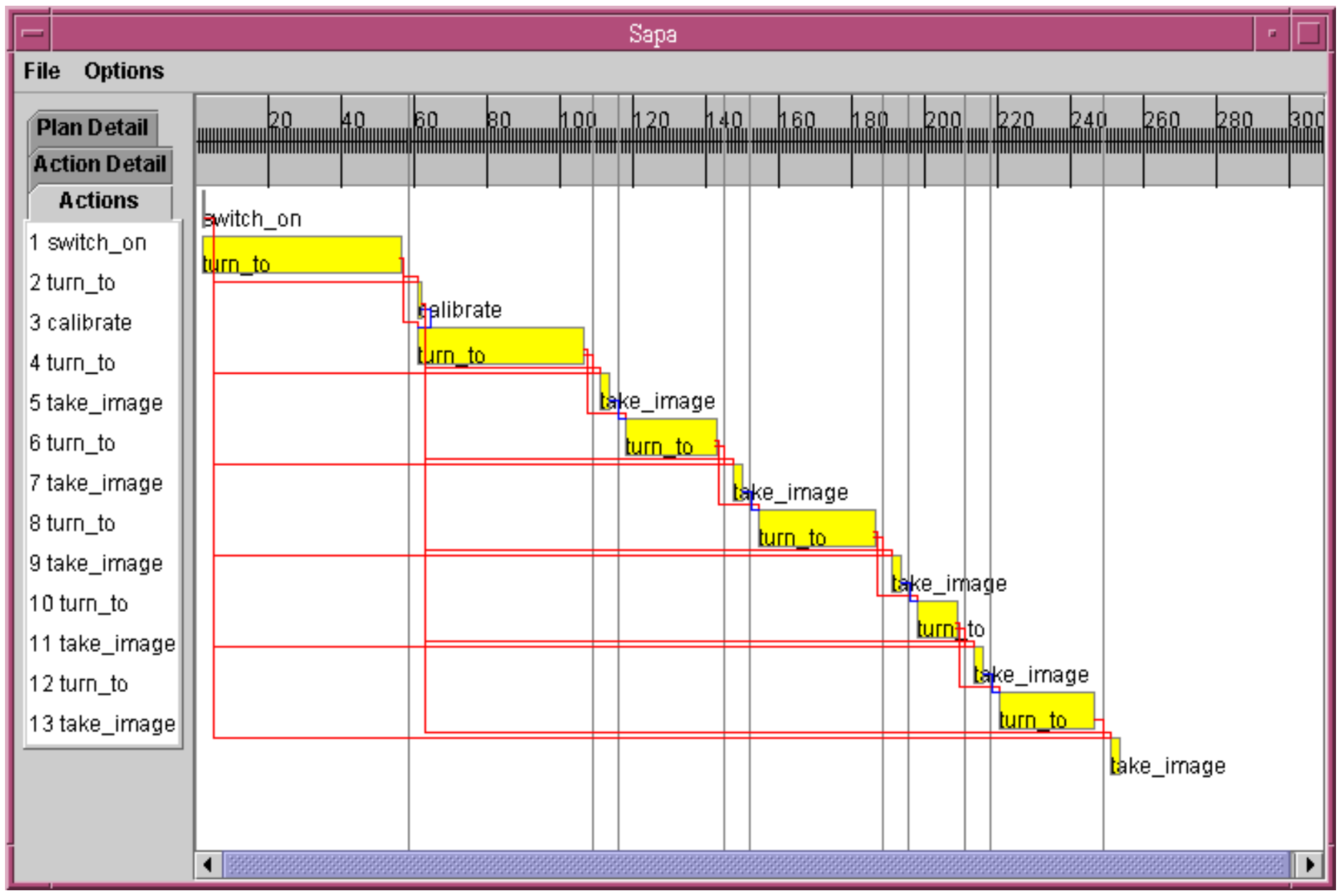

Figure 11: Screen shot of *Sapa*'s GUI: PERT chart showing the actions' starting times and the precedence orderings between them.

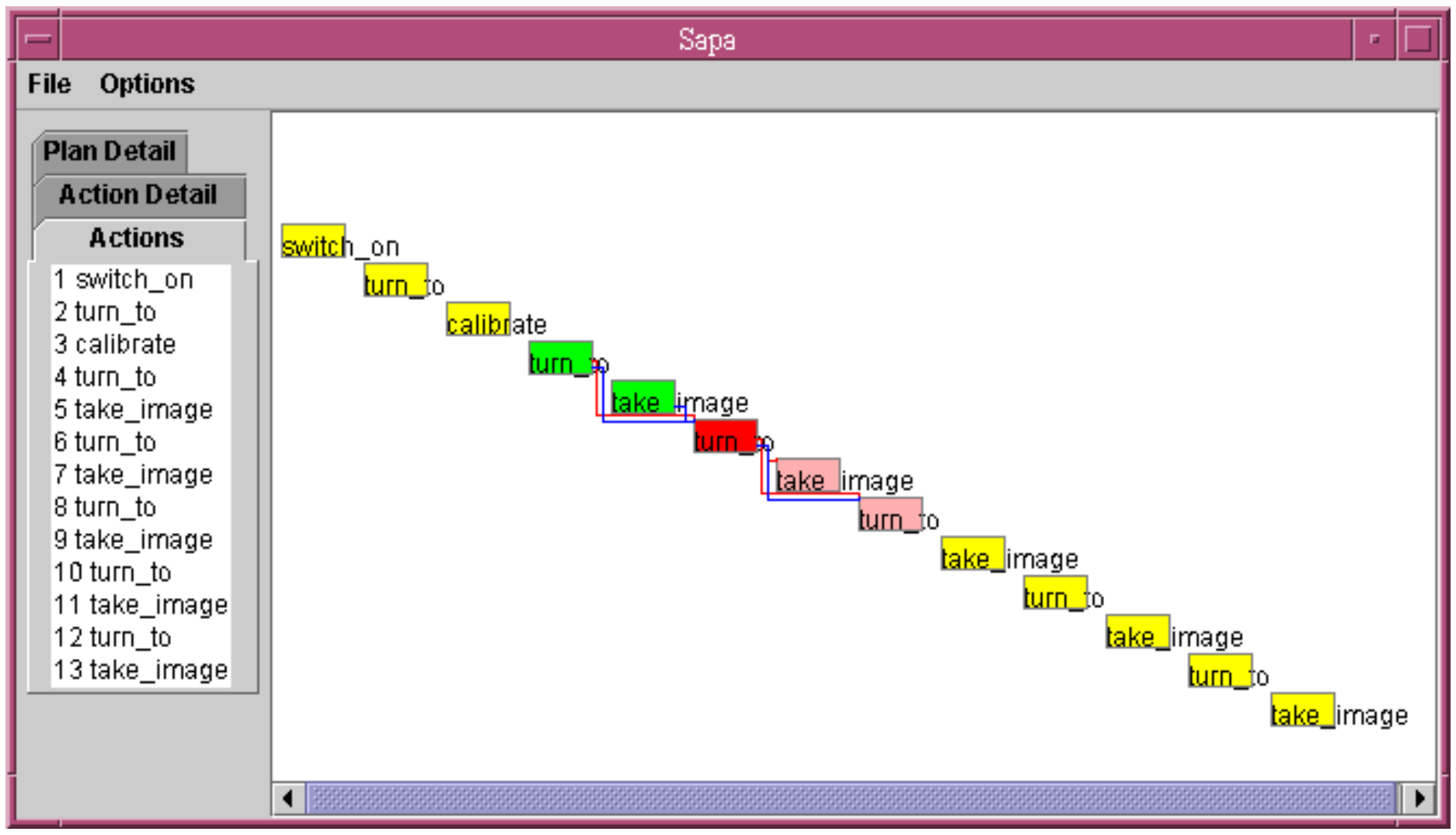

Figure 12: Screen shots of *Sapa*'s GUI: Gant chart showing different logical relations between a given action and other actions in the plan.

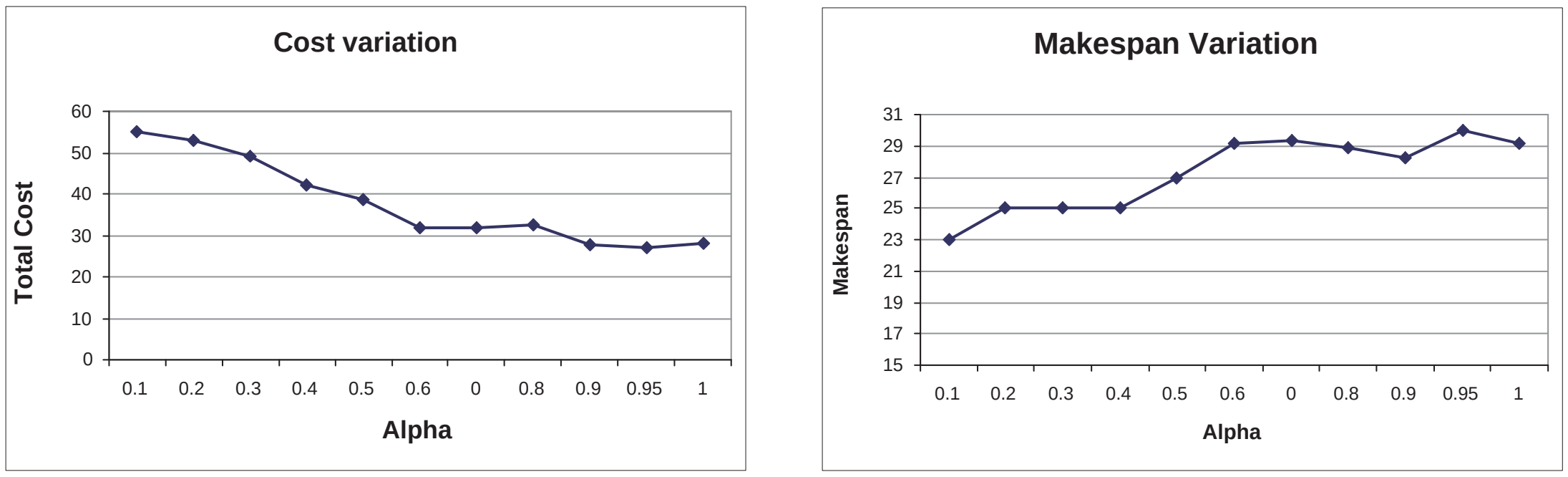

Figure 13: Cost and makespan variations according to different weights given to them in the objective function. Each point in the graph corresponds to an average value over 20 problems.

Our implementation is publicly available through the *Sapa* homepage[13]. Since the planner as well as the GUI are in JAVA, we also provide web-based interactive access to the planner.

## 8. Empirical Evaluation

We have subjected the individual components of the *Sapa* implementation to systematic empirical evaluation (c.f. Do & Kambhampati, 2001, 2002, 2003). In this section, we will describe the experiments that we conducted (Do & Kambhampati, 2001) to show that *Sapa* is capable of satisfying a variety of cost/makespan tradeoffs. Moreover, we also provide results to show the effectiveness of the heuristic adjustment techniques, the utility of different termination criteria, and the utility of the post-processing. Comparison of *Sapa* with other systems in the International Planning Competition is provided in the next section.

### 8.1 Component Evaluation

Our first test suite for the experiments, used to test *Sapa*'s ability to produce solutions with tradeoffs between time and cost quality, consisted of a set of randomly generated temporal logistics problems provided by Haslum and Geffner (2001). In this set of problems, we need to move packages between locations in different cities. There are multiple ways to move packages, and each option has different time and cost requirements. Airplanes are used to move packages between airports in different cities. Moving by airplanes takes only 3.0 time units, but is expensive, and costs 15.0 cost units. Moving packages by trucks between locations in different cities costs only 4.0 cost units, but takes a longer time of 12.0 time units. We can also move packages between locations inside the same city (e.g. between offices and airports). Driving between locations in the same city costs 2.0 units and takes 2.0 time units. Loading/unloading packages into a truck or airplane takes 1.0 unit of time and costs 1.0 unit.

We tested the first 20 problems in the set with the objective function specified as a linear combination of both total execution cost and makespan values of the plan. Specifically, the objective

12. The GUI was developed by Dan Bryce
13. http://rakaposhi.eas.asu.edu/sapa.html

function was set to

$$O = \alpha.C(Plan) + (1 - \alpha).T(Plan)$$

We tested with different values of $\alpha : 0 \leq \alpha \leq 1$. Among the techniques discussed in this paper, we used the sum-propagation rule, infinite look-ahead, and relax-plan extraction using static mutex relations. Figure 13 shows how the average cost and makespan values of the solution change according to the variation of the $\alpha$ value. The results show that the total execution cost of the solution decreases as we increase the $\alpha$ value (thus, giving more weight to the execution cost in the overall objective function). In contrast, when $\alpha$ decreases, giving more weight to makespan, the final cost of the solution increases and the makespan value decreases. The results show that our approach indeed produces solutions that are sensitive to an objective function that involves both time and cost. For all the combinations of $\{problem, \alpha\}$, 79% (173/220) are solvable within our cutoff time limit of 300 seconds. The average solution time is 19.99 seconds and 78.61% of the instances can be solved within 10 seconds.

#### 8.1.1 EVALUATION OF DIFFERENT TERMINATION CRITERIA

Figure 14 shows the comparison results for zero, one, and infinite lookahead for the set of metric temporal planning problems in the competition. We take the first 12 problems in each of the four temporal domains: ZenoTravel-Time, DriverLog-Time, Satellite-Time, and RoversTime. We set $\alpha = 1$ in the objective function, making it entirely cost-based. The action costs are set to 1 unit. As discussed in Section 3.3, zero-lookahead stops the cost propagation process at the time point where there is a solution under the relaxed condition. K-lookahead spends extra effort to go beyond that time point in hope of finding better quality (relaxed) solution to use as heuristic values to guide the search. The running condition is specified in the caption of the figure.

For most of the problems in the three domains ZenoTravel-Time, DriverLog-Time, and Satellite-Time, infinite-lookahead returns better quality solutions in shorter time than one-lookahead, which in turn is generally better than zero-lookahead. The exception is the Rovers-Time domain, in which there is virtually no difference in running time or solution quality between the different look-ahead options.

The following is a more elaborate summary of the results in Figure 14. The top three figures show the running time, cost, and makespan comparisons in the ZenoTravel domain (*Time* setting). In this domain, within the time and memory limit, infinite-lookahead helps to solve 3 more problems than one-lookahead and 2 more than zero-lookahead. In all but one problem (problem 10), infinite-lookahead returns equal (three) or better (eight) cost solution than zero-lookahead. Compared to one-lookahead, it's better in five problems and equal in six others. For the makespan value, infinite-lookahead is generally better, but not as consistent as other criteria. The next three lower figures show the comparison results for the DriverLog-Time domain. In this domain, infinite and one-lookahead solve one more problem than zero-lookahead, infinite-lookahead is also faster than the other two options in all but one problem. The costs (number of actions) of solutions returned by infinite-lookahead are also better than all but two of the problems (in which all three solutions are the same). One-lookahead is also equal to or better than zero-lookahead in all but two problems. In the Satellite-Time domain, while infinite and one-lookahead solve three more (of twelve) problems than zero-lookahead, there is no option that consistently solves problems faster than the other. However, the solution quality (cost) of infinite and one-lookahead is consistently better than zero-lookahead. Moreover, the solution cost of plans returned by infinite-lookahead is worse than one-lookahead in only one problem while being slightly better in 6 problems. In this domain, it seems that there is big improvement from zero to one look-ahead, while infinite-lookahead is a slight improvement over one-lookahead. (The plots for the Rovers domain are not shown in the figure as all the different look-ahead options seem to lead to near identical results in that domain.) Finally, since the heuristic is based completely on cost ($\alpha$ =1), we do not, in theory, expect to see any conclusive patterns in the makespan values of the solutions produced for the different lookahead options.

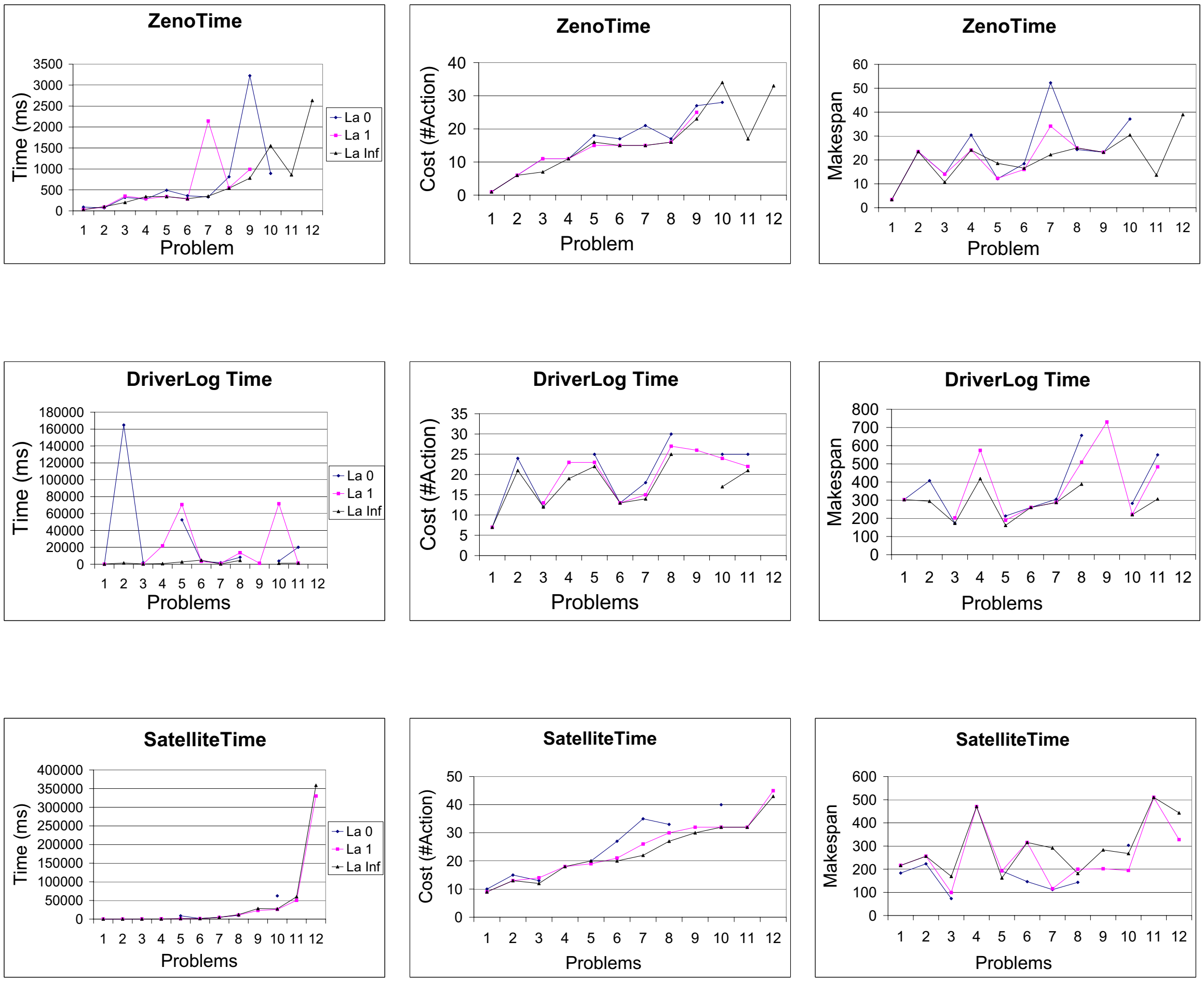

Figure 14: Comparison of the different lookahead options in the competition domains. These experiments were run on a Pentium III-750 WindowsXP machine with 256MB of RAM. The time cutoff is 600 seconds.

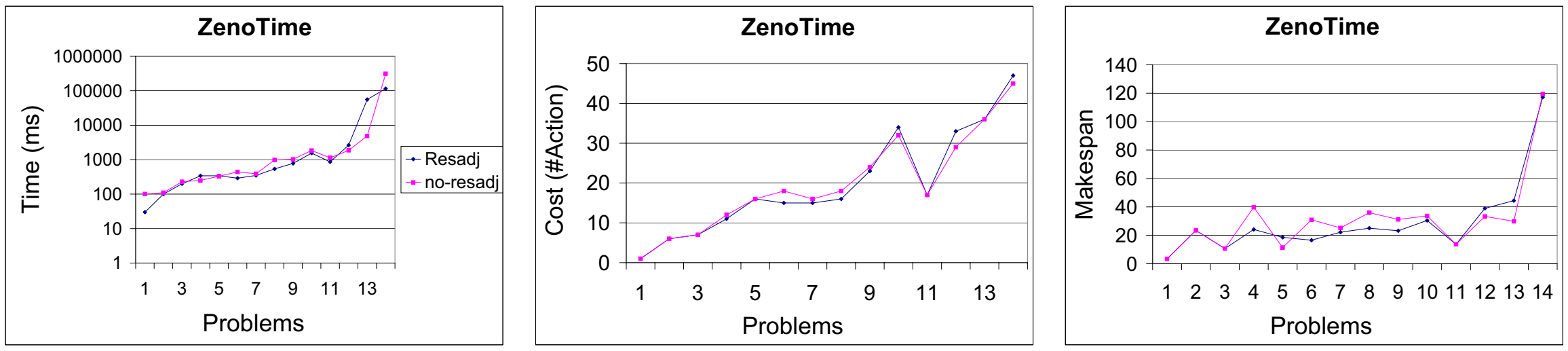

Figure 15: Utility of the resource adjustment technique on ZenoTravel (time setting) domain in the competition. Experiments were run on a WindowsXP Pentium III 750MHz with 256MB of RAM. Time cutoff is 600 seconds.

### 8.1.2 UTILITY OF THE RESOURCE ADJUSTMENT TECHNIQUE

In our previous work (Do & Kambhampati, 2001), we show that the resource adjustment technique can lead to significant quality and search speed improvements in problems such as the metric temporal logistics domain in which there are several types of resource consumption objects like trucks and airplanes.

In the competition, there are two domains in which we can test the utility of the resource adjustment technique discussed in Section 5.2. The ZenoTravel domain and the Rovers domain have actions consuming resources and other (refueling) actions to renew them. Of these, the resource adjustment technique gives mixed results in the ZenoTravel domain and has no effect in the Rovers domain. Therefore, we only show the comparison for the ZenoTravel domain in Figure 15. In the ZenoTravel domain, *Sapa* with the resource adjustment runs slightly faster in 10 of 14 problems, returns shorter solutions in 5 problems and longer solutions in 3 problems. The solution makespan with the resource adjustment technique is also generally better than without the adjustment technique. In conclusion, the resource adjustment technique indeed helps *Sapa* in this domain. In contrast, in the Rovers domain, this technique is of virtually no help. Actually, in the Rovers domain, the number of search nodes with and without the resource adjustment is the same for all solved problems. One reason maybe that in the Rovers domain, there are additional constraints on the *recharge* action so that it can only be carried out at a certain location. Therefore, even if we know that we need to add a recharge action to the current relaxed plan, we may not be able to add it because the plan does not visit the right location.

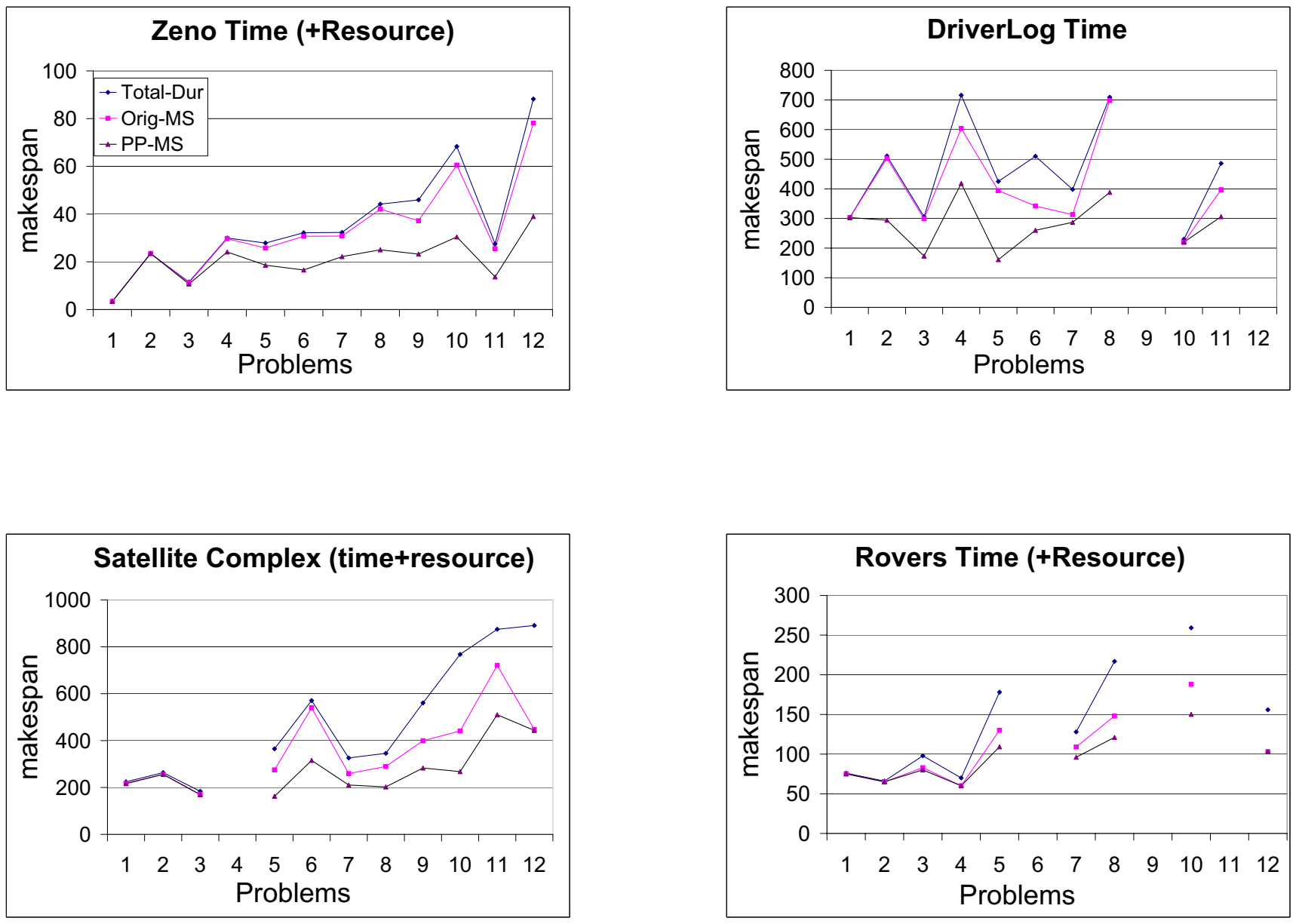

Figure 16: Utility of the greedy post-processing approach for problems in domains ZenoTravel-Time, DriverLog-Time, Satellite-Complex, and Rovers-Time in the IPC3.

#### 8.1.3 UTILITY OF POST-PROCESSING P.C. PLANS TO O.C. PLANS

Figure 16 shows the utility of the greedy post-processing technique discussed in Section 6. The test suite contains the same set of problems used in the first test, which are the first 12 problems in the ZenoTravel-Time, DriverLog-Time, Satellite-Complex, and Rovers-Time domain. The graphs in Figure 16 show the comparisons of makespan values of (i) total duration of all actions in the plan (makespan of a serial plan); (ii) original parallel position-constrained (p.c) plans returned by *Sapa*, and (iii) order-constrained (o.c) plans returned after post-processing. The graphs show that the greedy post-processing approach helps improving the makespan values in all domains. On average, it improves the makespan values of the original plans by 32.4% in the ZenoTravel-Time domain, 27.7% in the DriverLog-Time domain, 20.3% in the Satellite-Complex domain, and 8.7% in the RoversTime domain. Compared to the serial plans, the greedily partialized o.c plans improved the makespan values 24.7%-38.9%.

The CPU times for greedy partialization are very small. Specifically, they were less than 0.1 seconds for all problems with the number of actions ranging from 1 to 68. Thus, using our partialization algorithm as a post-processing stage essentially preserves the significant efficiency advantages of position constrained planners such as *Sapa*, GRT and MIPS, that search in the space of p.c. plans, while improving the temporal flexibility of the plans generated by those planners.

In our related work (Do & Kambhampati, 2003), we present additional results for the SimpleTime setting of those domains and do a comparison with an optimal post-processing technique discussed in the same paper.

### 8.2 *Sapa* in the 2002 International Planning Competition

We entered an implementation of *Sapa*, using several of the techniques discussed in this paper, in the recent International Planning Competition. The specific techniques used in the competition version of *Sapa* are infinite look-ahead termination of cost propagation (Section 3.3), resource adjustment (Section 5.2), and greedy post-processing (Section 6). In the competition, we focused solely on the metric/temporal domains.

The sophisticated support for multi-objective search provided by *Sapa* was not fully exploited in the competition, since action cost is not part of the standard PDDL2.1 language used in the competition.[14] The competition did evaluate the quality of solution plans both in terms of number of actions and in terms of the overall makespan. Given this state of affairs, we assumed unit cost for all actions, and ran *Sapa* with $\alpha = 1$, thus making the search sensitive only to the action costs. Infinite-lookahead was used for cost propagation. This strategy biased *Sapa* to produce low cost plans (in terms of number of actions). Although the search was not sensitive to makespan optimization, the greedy post processing of p.c. plans to o.c. plans improved the makespan of the solutions enough to make *Sapa* one of the best planners in terms of the overall quality of solutions produced.[15]

The competition results were collected and distributed by the IPC3's organizers and can be found at the competition website (Fox & Long, 2002). Detailed descriptions of domains used in the competition are also available at the same place. The temporal planning domains in the competition came in two sets, one containing two domains, Satellite and Rovers (adapted from NASA domains), and the other containing three domains: *Depots*, *DriverLogistics* and *Zeno Travel*. In the planning competition, each domain had multiple versions–depending on whether or not the actions had durations and whether actions use continuous resources. *Sapa* participated in the highest levels of PDDL2.1 (in terms of the complexity of temporal and metric constraints) for the five domains listed above.

Figures 17 and 18 show that five planners (*Sapa*, LPG, MIPS, TP4, and TPSYS) submitted results for the *timed* setting and only three (*Sapa*, MIPS, and TP4) were able to handle the *complex* setting of the Satellite domain. In the *timed* setting, action durations depend on the setting of instruments aboard a particular satellite and the direction it needs to turn to. The *"complex"* setting is further complicated by the fact that each satellite has a different capacity limitation so that it can only store a certain amount of image data. Goals involve taking images of different planets and stars located at different coordinate directions. To achieve the goals, the satellite equipped with the right set of instruments should turn to the right direction, calibrate and take the image.

For the *timed* setting of this Satellite domain, Figure 17 shows that among the five planners, *Sapa*, LPG and MIPS were able to solve 19 of 20 problems while TP4 and TPSYS were able to solve 2 and 3 problems respectively. For quality comparison, LPG with the *quality* setting was able to return solutions with the best quality, *Sapa* was slightly better than LPG with the *speed* setting and was much better than MIPS. LPG with the *speed* setting is generally the fastest, followed by MIPS and then *Sapa* and LPG with the *quality* setting. For the *complex* setting, Figure 18 shows that, among the three planners, *Sapa* was able to solve the most problems (16), and generated plans of higher quality than MIPS. TP4 produced the highest quality solutions, but was able to solve only

14. Some of the competition problems did have explicit objective functions, and in theory, we could have inferred the action costs from these objective functions (see the discussion in Section 4.3). We have not yet done this, given that the "plan metric" field of PDDL2.1 had not been fully standardized at the time of the competition.
15. To be sure, makespan optimal planners such as TP4 can produce much shorter plans–but their search was so inefficient that they were unable to solve most problems.

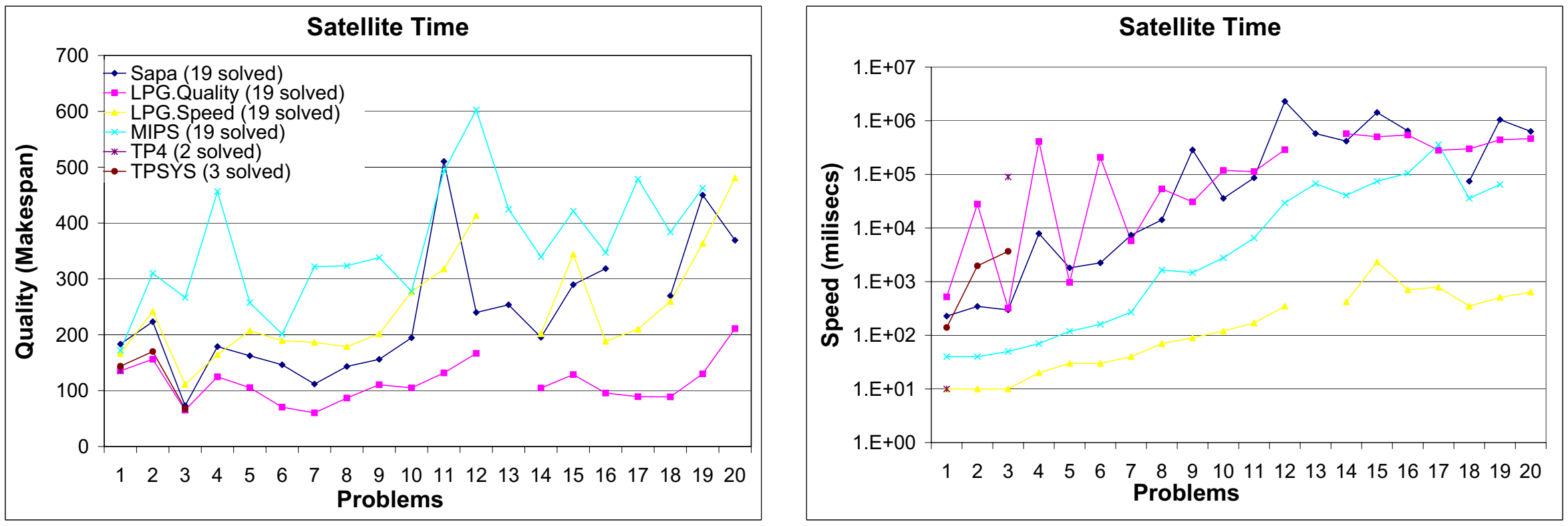

Figure 17: Results for the *time* setting of the Satellite domain (from IPC3 results).

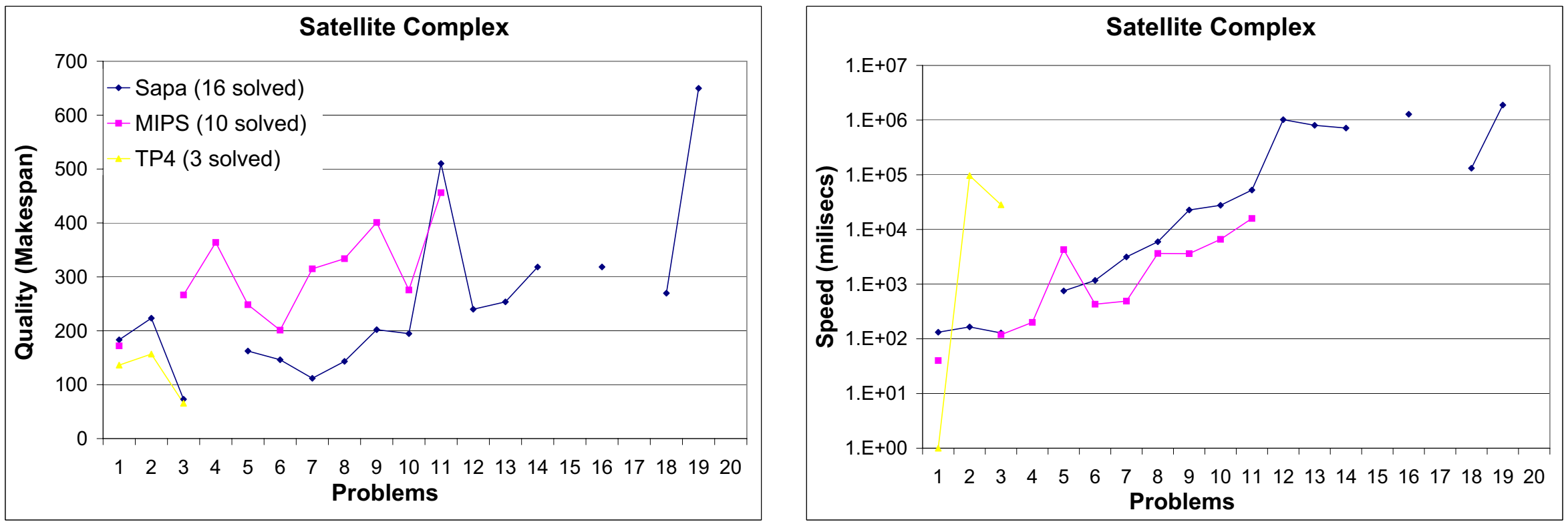

Figure 18: Results for the *complex* setting of the Satellite domain (from IPC3 results).

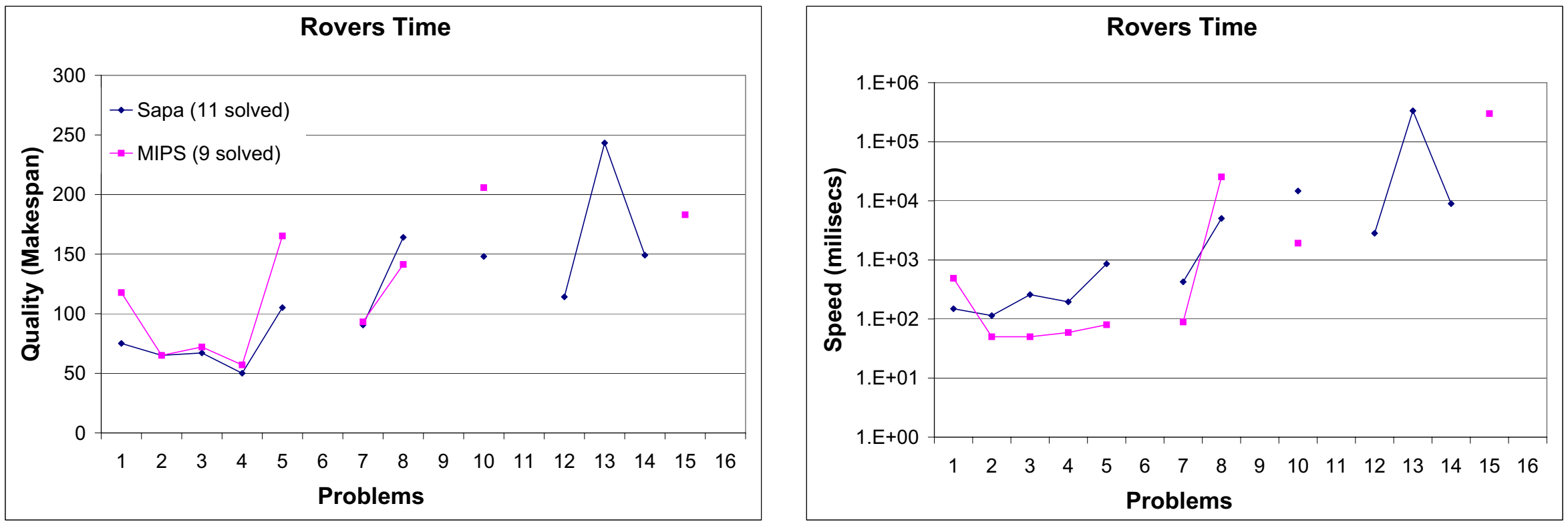

Figure 19: Results for the *time* setting of the Rover domain (from IPC3 results).

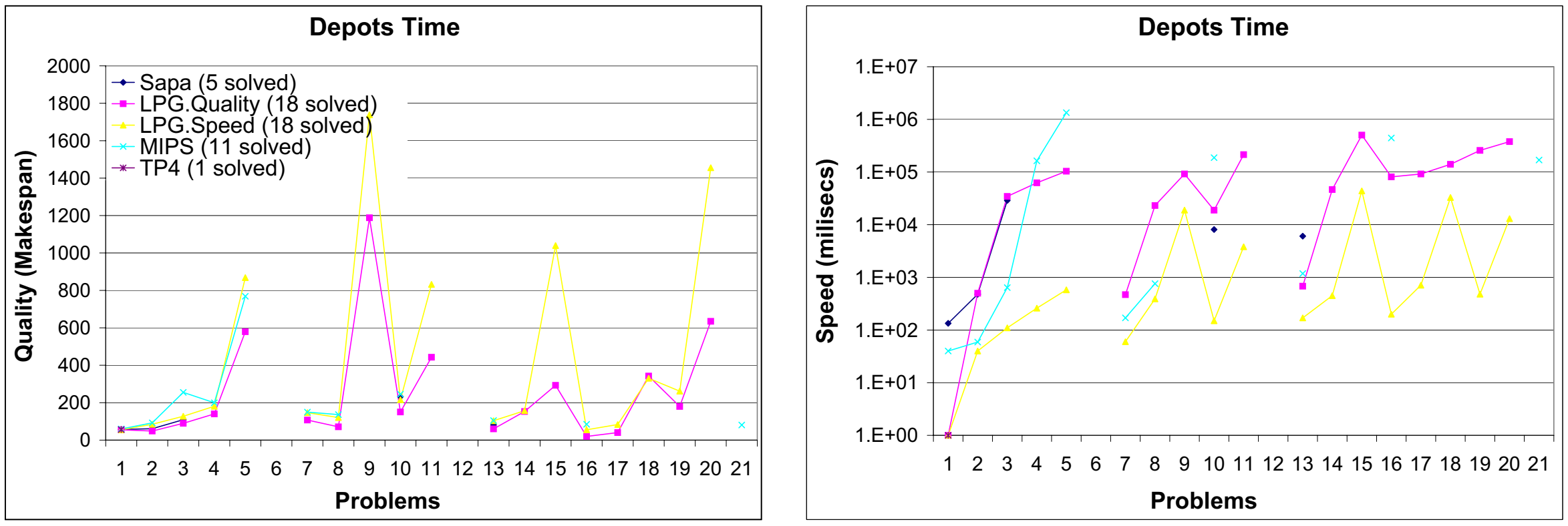

Figure 20: Results for the *time* setting of the Depots domain (from IPC3 results).

the three smallest problems. The solving times of *Sapa* are slightly higher than MIPS, but are much better than TP4.

The "timed" version of the Rover domain requires that a set of scientific analyses be done using a number of rovers. Each rover carries different equipment, and has a different energy capacity. Moreover, each rover can only recharge its battery at certain points that are under the sun (which may be unreachable). Figure 19 shows that only *Sapa* and MIPS were able to handle the constraints in this problem set. *Sapa* again solved more problems (11 vs. 9) than MIPS and also returned better or equal quality solutions in all but one case. The solving time of MIPS is better than *Sapa* in 6 of 9 problems where it returns the solutions.

In the second set of problems, which come with temporal constraints, there are three domains: *Depots*, *DriverLogistics* and *Zeno Travel*. *Sapa* participated at the highest level, which is the *“timed”* settings for these three domains. Figure 20 shows the comparison between *Sapa* and three other planners (LPG, MIPS, and TP4) that submitted results in the *Depots* domain. In this domain, we need to move crates (packages) between different places. The loading actions that place the crates into each truck are complicated by the fact that they need an *empty* hoist. Thus, the *Depot* domain looks like a combination of the original logistics and blockworlds domains. Drive action durations depend on the distances between locations and the speed of the trucks. Time for loading the crates depends on the power of the hoist that we use. There is no resource consumption in this highest level. In this domain, *Sapa* was only able to solve five problems, compared to 11 by MIPS and 18 by LPG. TP4 solved only one problem. For the five problems that *Sapa* was able to solve, the solution quality is as good as other planners. For the speed comparison, LPG with *speed* setting is clearly faster than the other planners. We speculate that the poor performance of *Sapa* in this domain is related to two factors: (i) negative interactions between subgoals, largely ignored by *Sapa*, are an important consideration in this domain and (ii) the number of ground actions in this domain is particularly high, making the computation of the planning graph quite costly.

Figure 21 shows how *Sapa* performance compares with other planners in the competition on the *time* setting of the *DriveLog* domain. This is a variation of the original *Logistics* domain in which trucks rather than airplanes move packages between different cities. However, each truck requires a driver, so a driver must walk to and board a truck before it can move. Like the *Depot* domain, there is no resource consumption. The durations for walking and driving depend on the specified *time-to-walk* and *time-to-drive*. In this domain, *Sapa* solved 14 problems compared to 20 by LPG, 16 by MIPS and 1 by TP4. The quality of the solutions by different planners are very similar. For the speed comparison, LPG with *speed* setting is fastest, then MIPS, then *Sapa* and LPG with *quality* setting.

Finally, Figure 22 shows the performance of *Sapa* in the ZenoTravel domain with *time* setting. In this domain, passengers travel between different cities by airplanes. The airplanes can choose to fly with different speeds (fast/slow), which consume different amounts of fuel. Airplanes have different fuel capacity and need to refuel if they do not have enough for each trip. In this domain, *Sapa* and LPG solved 16 problems while MIPS solved 20. The solution quality of *Sapa* and MIPS are similar and in general better than LPG with either *speed* or *quality* setting. LPG with *speed* setting and MIPS solved problems in this domain faster than *Sapa* which is in turn faster than LPG with *quality* setting.

In summary, for problems involving both metric and temporal constraints in IPC3, *Sapa* is competitive with other planners such as LPG or MIPS. In particular, *Sapa* solved the most problems and returned the plans with best solution quality in the highest setting for the two domains *Satellite* and *Rovers*, which are adapted from NASA domains. A more detailed analysis of the competition results is presented by Long and Fox (2003).

## 9. Related Work and Discussion

Although there have been several recent domain-independent heuristic planners aimed at temporal domains, most of them have been aimed at makespan optimization, ignoring the cost aspects. For example, both TGP (Smith & Weld, 1999) as well as TP4 (Haslum & Geffner, 2001) focus on makespan optimization and ignore the cost aspects of the plan. As we have argued in this paper, ultimately, metric temporal planners need to deal with objective functions that are based on both makespan and cost. One recent research effort that recognizes the multi-objective nature of planning is the MO-GRT system (Refanidis & Vlahavas, 2001a). On one hand, the MO-GRT approach is more general than our approach in the sense that it deals with a set of non-combinable quality metrics. The MO-GRT approach however treats time similar to other consumable resources (with infinite capacity). Temporal constraints on the planning problems (such as when an effect should occur during the course of action), goal deadlines, or concurrency are *ignored* in order to scale down the problem to the classical planning assumptions. Metric-FF (Hoffmann, 2002) and MIPS (Edelkamp, 2001) are two other forward state space planners that can handle resource constraints. Both of them generate sequential plans. MIPS handles durative actions by putting in the action duration and post-processing the sequential p.c plans. Multi-Pegg (Zimmerman, 2002) is another recent planner that considers cost-time tradeoffs in plan generation. Multi-Pegg is based on the Graphplan approach, and focuses on classical planning problems with non-uniform cost actions. ASPEN (Chien et al., 2000) is another planner that recognizes the multi-attribute nature of plan quality. ASPEN advocates an iterative repair approach for planning, that assumes the availability of a variety of hand-coded plan repair strategies and their characterization in terms of their effects on the various dimensions of plan quality. LPG (Gerevini & Serina, 2002) is another planner that employs local search techniques over the action-graph. Unlike ASPEN, LPG considers domain independent repair strategies that involve planning graph-based modifications.

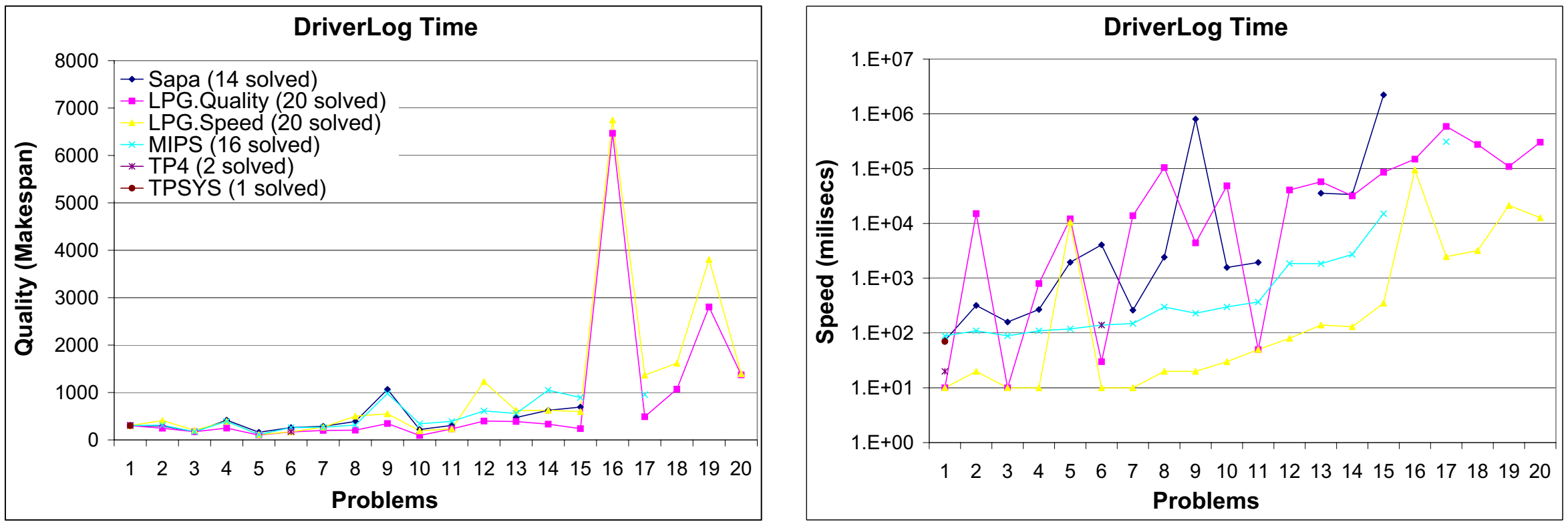

Figure 21: Results for the *time* setting of the DriverLog domain (from IPC3 results).

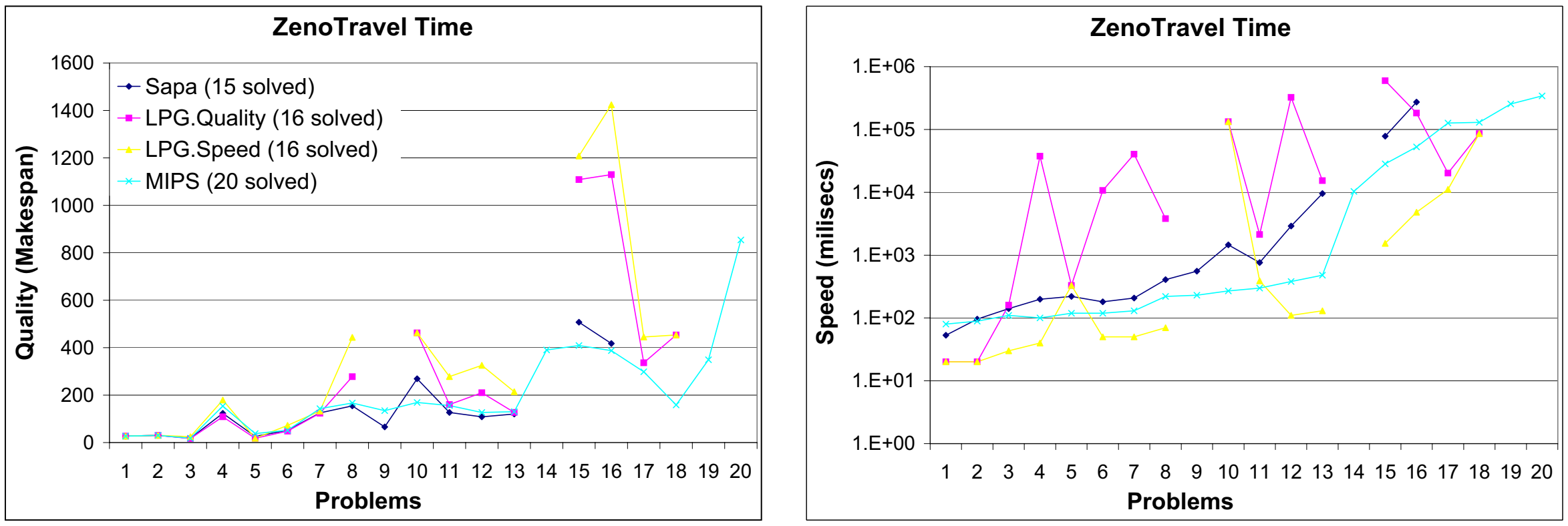

Figure 22: Results for the *time* setting of the ZenoTravel domain (from IPC3 results).

Although we evaluated our cost-sensitive heuristics in the context of *Sapa*, a forward chaining planner, the heuristics themselves can also be used in other types of planning algorithms. For example, TGP can be made cost-sensitive by making it propagate the cost functions as part of planning graph expansion. These cost functions can then be used as a basis for variable and value ordering heuristics to guide its backward branch-and-bound search. A similar approach in classical planning has been shown to be successful by Kambhampati and Nigenda (2000).

Besides Graphplan-based approaches, our framework can also be used in both forward and backward state-space and partial order planners to guide the planning search. It is possible due to the fact that directional searches (forward/backward) all need to evaluate the distances between an initial state and a set of temporal goals.

Our work is also related to other approaches that use planning graphs as the basis for deriving heuristic estimates such as Graphplan-HSP (Kambhampati & Nigenda, 2000), AltAlt (Nguyen et al., 2001), RePOP (Nguyen & Kambhampati, 2001), and FF (Hoffmann & Nebel, 2001). In the context of these efforts, our contribution can be seen as providing a way to track cost as a function of time on planning graphs. An interesting observation is that cost propagation is in some ways inherently more complex than makespan propagation. For example, once a set of literals enter the planning graph (and are not mutually exclusive), the estimate of the makespan of the shortest plan for achieving them does not change as we continue to expand the planning graph. In contrast, the estimate of the cost of the cheapest plan for achieving them can change until the planning graph levels off. This is why we needed to carefully consider the effect of different criteria for stopping the expansion of the planning graph on the accuracy of the cost estimates (Section 3.3).

Another interesting point is that within classical planning, there was often a confusion between the notions of cost and makespan. For example, the “length of a plan in terms of number of actions” can either be seen as a cost measure (if we assume that actions have unit costs), or a makespan measure (if we assume that the actions have unit durations). These notions get teased apart naturally in metric temporal domains.

In this paper, we concentrated on developing heuristics that can be sensitive to multiple dimensions of plan quality (specifically, makespan and cost). An orthogonal issue in planning with multiple criteria is how the various dimensions of plan quality should be combined during optimization. The particular approach we adopted in our empirical evaluation–namely, considering a linear combination of cost and time–is by no means the only reasonable way. Other approaches involve non-linear combinations of the quality criteria, as well as "tiered" objective functions (e.g. rank plans in terms of makespan, breaking ties using cost). A related issue is how to help the user decide the "weights" or "tiers" of the different criteria. Often the users may not be able to articulate their preferences between the various quality dimensions in terms of precise weights. A more standard way out of this dilemma involves generating all non-dominated or Pareto-optimal plans[16], and presenting them to the user. Unfortunately, often the set of non-dominated plans can be exponential (c.f., Papadimitriou & Yannakakis, 2001). The users are then expected to pick the plan that is most palatable to them. Unfortunately, the users may not actually be able to judge the relative desirability of plans when the problems are complex and the plans are long. Thus, a more practical approach may involve resorting to other indirect methods such as preference elicitation techniques (c.f. Chajewska, Getoor, Norman, & Shahar., 1998).

## 10. Conclusion

In this paper, we presented *Sapa*, a domain-independent heuristic forward chaining planner that can handle durative actions, metric resource constraints, and deadline goals. *Sapa* is a forward-chaining planner that searches in the space of time-stamped states. It is designed to be capable of handling the multi-objective nature of metric temporal planning. Our technical contributions include (i) a planning-graph based method for deriving heuristics that are sensitive to both cost and makespan (ii) an easy way of adjusting the heuristic estimates to take the metric resource limitations into account and (iii) a way of post-processing the solution plans to improve their execution flexibility. We described the technical details of extracting the heuristics and presented an empirical evaluation of the current implementation of *Sapa*. An implementation of *Sapa* using a subset of the techniques presented in this paper was one of the best domain independent planners for domains with metric and temporal constraints in the third International Planning Competition, held at AIPS-02.

We are extending *Sapa* in several different directions. To begin with, we want to make *Sapa* support more expressive domains, including exogenous events and a richer set of temporal and resource constraints (e.g a rover can not recharge the battery after sunset). Another direction involves extending our multi-objective search to involve other quality metrics. While we considered cost of a plan in terms of a single monetary cost associated with each action, in more complex domains, the cost may be better defined as a vector comprising the different types of resource consumption. Further, in addition to cost and makespan, we may also be interested in other measures of plan quality such as robustness and execution flexibility of the plan. Our longer term goal is to support plan generation that is sensitive to this extended set of tradeoffs. To this end, we plan to extend our methodology to derive heuristics sensitive to a larger variety of quality measures. Finally, we also plan to consider the issues of planner-scheduler interactions in the context of *Sapa* (c.f., Srivastava, Kambhampati, & Do, 2001).

16. A plan $P$ is said to be dominated by $P'$ if the quality of $P'$ is strictly superior to that of $P$ in at least one dimension, and is better or equal in all other dimensions (Dasgupta, Chakrabarti, & DeSarkar., 2001; Papadimitriou & Yannakakis, 2001).

## Acknowledgments

We thank Daniel Bryce for developing the GUI for *Sapa.* We specially thank David E. Smith for his many insightful and detailed comments on the paper. We also thank the JAIR reviewers for their very helpful comments. This research is supported in part by the NSF grant IRI-9801676, and the NASA grants NAG2-1461 and NCC-1225.